%% file: main.tex
\PassOptionsToPackage{table}{xcolor}
\documentclass[11pt]{tea}
\microtypesetup{expansion=false}

\usepackage[utf8]{inputenc}
\usepackage[T1]{fontenc}
\usepackage{amsmath}
\usepackage{nicefrac}
\usepackage{url}
\usepackage{enumitem}
\usepackage{algorithm}
\usepackage{algorithmic}
\usepackage{tablefootnote}
\usepackage{pifont}
\usepackage{fvextra}
\usepackage{makecell}
\usepackage{bbm}
\usepackage{threeparttable}
\usepackage{wrapfig}
\usepackage{arydshln}
\usepackage{xspace}
\usepackage{soul}

\newcommand{\verimap}{\texttt{VeriMap}\xspace}
\newcommand{\ours}{\texttt{MirageCheck}\xspace}

\newcommand{\answerTODO}[1][]{\textcolor{red}{\bf [TODO]}}

\definecolor{insightbg}{HTML}{F4FAF4}
\definecolor{insightborder}{HTML}{1F2D30}

\newtcolorbox[auto counter]{insightbox}{%
  enhanced,
  colback=insightbg,
  colframe=insightborder,
  boxrule=1.2pt,
  arc=4pt,
  outer arc=4pt,
  boxsep=0pt,
  left=14pt,
  right=14pt,
  top=10pt,
  bottom=10pt,
  before skip=10pt,
  after skip=10pt,
  width=\linewidth,
  fontupper=\normalfont
}

\newtcolorbox{PromptBox}[1]{
  colback=gray!5!white,
  colframe=gray!75!black,
  fonttitle=\bfseries,
  title=#1,
  arc=4pt,
  outer arc=4pt,
  breakable,
  left=6pt, right=6pt
}

\title{
Verification Mirage: Mapping and Repairing the Reliability Boundary of Self-Verification in Medical VQA}

\author[12]{Ruinan Jin}
\author[12]{Beidi Zhao}
\author[12]{Myeongkyun Kang}
\author[1]{Jiaqi Liu}
\author[4]{Behzad Bozorgtabar}
\author[3]{Qiong Zhang}
\author[12]{Xiaoxiao Li}

\affiliation[1]{The University of British Columbia}
\affiliation[2]{Vector Institute}
\affiliation[3]{Redmin University of China}
\affiliation[4]{Aarhus University}

\abstract{
\input{sections/abstract}
}

\begin{document}

\maketitle

\input{sections/introduction}
\input{sections/related_work}
\input{sections/method}
\input{sections/experiments}
\input{sections/conclusion}

\newpage
\input{sections/statements}

\bibliographystyle{plainnat}
\bibliography{references}

\clearpage
\newpage
\beginappendix
\input{sections/appendix}


\end{document}

%% file: sections/abstract.tex
Self-verification is increasingly used to improve the reliability of vision-language models (VLMs) and provide feedback for iterative self-correction, yet it is useful only when checking an answer provides information that discriminates correct from incorrect predictions rather than simply re-solving the same visual problem. We introduce \verimap{}, an auditing framework for mapping this reliability boundary in medical visual question answering, where many answers hinge on fine-grained but auditable visual distinctions. Across six VLMs, five datasets, and seven task types, \verimap{} reveals a \emph{verification mirage}: vanilla verifiers accept 71.0\% of incorrect candidates on average, and all six models are more likely to accept their own errors than errors from other models. When reused as feedback, these false acceptances can persist across self-correction rounds. We find that stronger access to the correct image or candidate-hidden re-solving is insufficient. Instead, verification becomes more selective when the candidate is reduced to a falsifiable visual distinction and the resulting observation governs the verdict. Guided by this finding, we introduce \emph{\ours{}}, which separates test construction, candidate-hidden visual measurement, and evidence-to-verdict binding through \textsc{Contrast}, \textsc{Measure}, and \textsc{Bind}. Across models and datasets, \ours{} reduces mean false acceptance by 14.6 percentage relative to the strongest candidate-independent baseline while improving balanced accuracy, and substantially reduces false acceptance during iterative correction. Together, our results show that reliable visual self-verification requires turning acceptance to discriminative evidence from the current image rather than another judgment of the same problem.

%% file: sections/introduction.tex
\section{Introduction}

Self-verification is often motivated by a familiar asymmetry: \emph{a proposed answer may be easier to check than to produce from scratch}~\citep{cobbe2021training,kadavath2022language,lightman2024let}. For an equation, deriving a solution may require several steps, while checking a proposed solution can require only substituting it back into the equation. Visual questions, however, do not automatically inherit this advantage. A candidate answer provides a hypothesis, but not necessarily the visual evidence needed to test it. When verification does not introduce a more discriminative check, the verifier may still need to recover the same finding, anatomy, count, or spatial relation from the image that was required during generation. Self-verification can therefore collapse into another judgment of nearly the same problem, now conditioned on a plausible answer. This limitation matters beyond a single verdict. In actor-verifier~\citep{gou2024critic}, self-correction~\citep{madaan2023self}, and self-improving systems~\citep{song2025mind}, verification is increasingly reused as feedback to revise an answer or guide subsequent behavior. If the verifier reproduces the actor's visual error, a local mistake can instead become a persistent positive signal.

This concern is especially sharp in medical VQA, where correctness often
hinges on fine-grained distinctions in findings, anatomy, laterality, counts,
or spatial relations~\citep{meddeb2025evaluating}. In our analysis in Appendix~\ref{app:medical_testbed}, false acceptance is consistently higher on sampled medical benchmarks than on general-domain benchmarks, by 22-34 percentage points across five VLMs. Many medical questions expose concrete, auditable answer-critical visual evidence.
This combination motivates our central question: \emph{\ul{when does medical visual self-verification provide a genuinely discriminative check rather than reproduce the original error?}}

Prior work suggests two natural explanations for failures of visual verification. \emph{First}, conditioning on a proposed answer can induce self-preference and agreement bias, motivating \emph{candidate-independent verification strategies}~\citep{xu2024pride,panickssery2024llm,wataoka2024self,andrade2025let}. \emph{Second}, multimodal models can produce judgments that are insufficiently grounded in the current image, motivating methods that \emph{stronger visual grounding} (e.g., hallucination)~\citep{lee2024volcano,liao2025can}. These two explanations lead to a natural question: \emph{are reducing candidate dependence and strengthening image dependence sufficient to make verification genuinely discriminative between correct and incorrect answers?}

\begin{figure}[t]
    \centering
    \includegraphics[width=\linewidth]{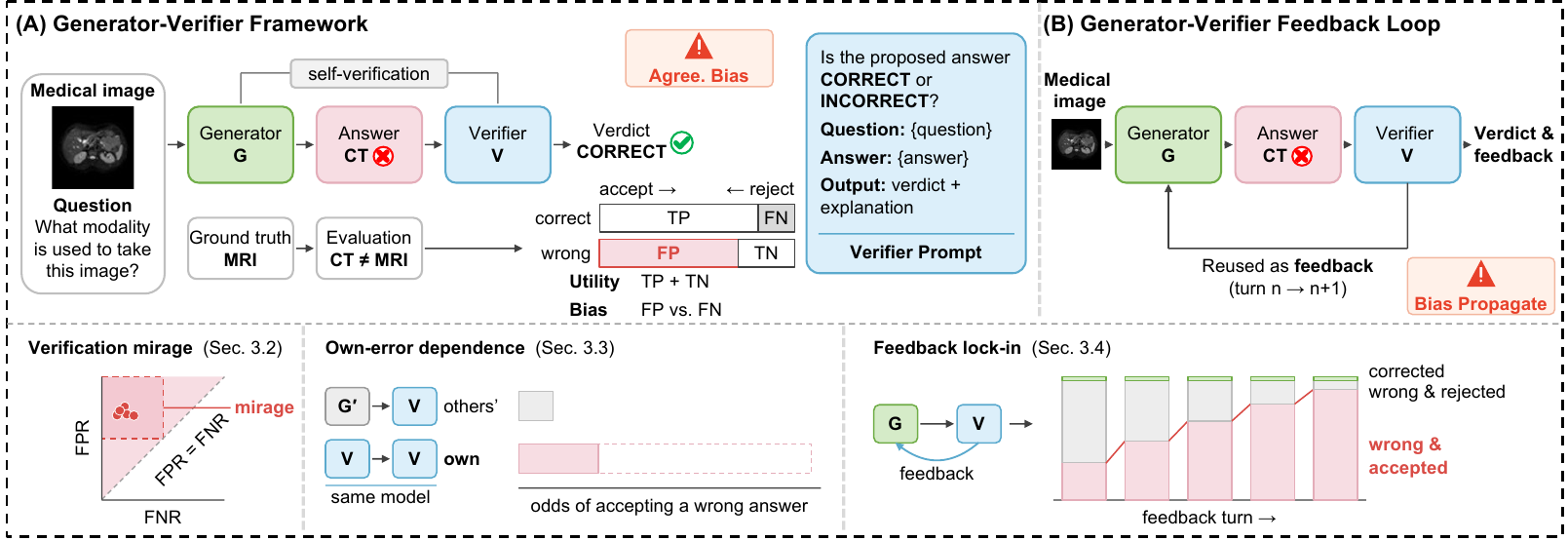}
    \caption{\textbf{Generator-verifier self-verification, feedback reuse, and the \verimap{} audit.}
\textbf{(A)} A generator answers a medical image question, and a verifier assesses the proposed answer. In the illustrated example, an incorrect answer is accepted, producing a false positive. The lower diagrams illustrate two findings from \verimap{}: the \emph{verification mirage}, where false acceptance exceeds false rejection, and \emph{own-error dependence}, where verifiers are more likely to accept their own incorrect answers than those from other models.
\textbf{(B)} The verifier's verdict and explanation are reused as feedback in subsequent turns. The lower diagram illustrates \emph{feedback lock-in}, where falsely accepted errors can persist across successive rounds of self-correction.}
    \label{fig:main}
\end{figure}

We study this question with \textbf{\verimap{}}, an auditing framework for mapping the reliability boundary of visual self-verification (Fig.~\ref{fig:main}). Across six VLMs, five medical VQA datasets, and seven task types, \verimap{} reveals a \textbf{verification mirage}: verifiers frequently accept incorrect candidates, and their failures are disproportionately concentrated on cases where generation has already gone wrong. Across the unified suite, verifiers accept 71.0\% of incorrect candidates on average. More importantly, this failure is not explained by general verifier permissiveness alone. In a complete cross-generator audit over wrong answers, all six verifiers preferentially accept errors produced by themselves even after controlling for verifier- and generator-specific effects. This matters when verifier outputs are reused as feedback, where the bias can propagate. In a five-turn actor-verifier self-improvement, false acceptance accumulates among initially wrong trajectories. We then test two natural explanations for the mirage (candidate independence and stronger visual grounding). Neither reliably separates correct from incorrect answers.

These results point to a more specific failure: the verifier may use the image and reason without seeing the candidate, yet still reproduce the same error. The missing part is answer-discriminative visual evidence. Instead, verification becomes more selective when it tests a concrete visual distinction between the candidate and an alternative, then bases the verdict
on what the image shows. In our frozen probe, this catches false acceptances more often than it overturns correct ones, motivating \ours{}.

Having localized this failure, we intervene directly on the missing visual check with \textbf{\ours{}}, a same-model protocol that replaces holistic re-judgment with an evidence-bound falsification test. \ours{} decomposes verification into three stages: \textsc{Contrast} identifies the answer-critical visual distinction that separates the candidate from plausible alternatives. \textsc{Measure} obtains the corresponding observation without revealing which hypothesis is the original answer, and \textsc{Bind} reconstructs the supported hypothesis from randomized alternatives using only the sealed evidence. The original acceptance is retained only when this reconstruction succeeds. \ours{} uses the same VLM checkpoint throughout and requires neither reference answers nor an external critic at inference time.

Our work makes three contributions. \textbf{(i) We introduce \verimap{}, a diagnostic framework for mapping the reliability boundary of medical visual self-verification.} Beyond aggregate verifier accuracy, \verimap{} separates false acceptance from discrimination, identifies where verifier failures concentrate relative to generation errors, delineates the boundary of this dependence, and exposes its consequence when verification is reused as feedback. \textbf{(ii) We localize a missing discriminative visual check.} Through controlled interventions and a falsification probe, we show that visual access and candidate independence alone are insufficient, and provide evidence that verification becomes more selective when the verdict is tied to answer-discriminative visual evidence.
\textbf{(iii) We translate this diagnosis into \ours{}, a targeted intervention for evidence-bound visual falsification.}
\ours{} replaces holistic re-judgment with \textsc{Contrast}, \textsc{Measure}, and \textsc{Bind}, improves verification across the five-dataset evaluation suite, and is further tested when its verdicts are reused for iterative self-correction.

%% file: sections/related_work.tex
\section{Related Work}

\paragraph{Self-verification, verifier reliability, and self-improvement.}
Self-verification supports selective generation, self-correction, and verifier-driven self-improvement across reasoning, code, and multimodal tasks
\citep{kadavath2022language,ren2023self,ni2023lever,chen2023teaching,madaan2023self,song2025mind}. Its usefulness, however, depends on the verifier providing information beyond the generator's original failure. Prior work
documents self-preference in LLM judges
\citep{xu2024pride,panickssery2024llm,wataoka2024self}, bias in multimodal verification \citep{andrade2025let}, and limits of same-model
verification for self-improvement \citep{song2025mind}.
We study a complementary question in visual verification. Rather than asking only whether verification is accurate or improves a solver, \verimap{} tests whether errors are preferentially accepted when produced by the verifier
itself, what makes verification discriminative, and how false acceptance propagates when verdicts are reused as feedback.

\paragraph{Multimodal verification and visual grounding.}
Multimodal self-correction and hallucination mitigation often improve reliability by eliciting candidate-independent judgments or strengthening
visual grounding~\citep{lee2024volcano,andrade2025let,sun2025mm,wu2025generate,ji2026causallens,yin2025clearsight}. These methods motivate two natural explanations for visual verification failure: candidate independence and stronger visual grounding. Our analysis separates these from a stricter requirement. A verifier can depend on the correct image and reason independently of the candidate while still
reproducing the same error. \verimap{} localizes the missing signal to answer-discriminative visual evidence, and \ours{} operationalizes this diagnosis by separating contrast construction, candidate-hidden measurement,
and evidence-to-verdict binding. We use medical VQA as an evidence-explicit testbed for this reliability boundary as detailed in Appendix~\ref{app:medical_testbed}.

%% file: sections/method.tex
\section{\verimap{}: Diagnosing Failures in Self-Verification}
\label{sec:verimap}

\verimap{} maps the reliability boundary of self-verification by asking whether verification discriminates correct from incorrect candidates, where its errors concentrate, and what happens when its verdicts are reused as feedback. We then test whether two prominent remedies, \emph{candidate independence} and \emph{stronger visual grounding}, are sufficient to resolve the failure and, if not, localize what additional signal is needed for a discriminative verdict.

\subsection{Evaluation Setup}
\label{sec:problem_formulation}

For each image-question pair $(I_i,q_i)$ with reference answer $a_i^*$, a VLM first generates a candidate answer
$\hat a_i = G(I_i,q_i)$. Comparing $\hat a_i$ with $a_i^*$ determines the candidate-correctness label $y_i\in\{0,1\}$, where $y_i=1$ indicates a correct candidate and $y_i=0$ an incorrect one. We then freeze $\hat a_i$ and ask a verifier (the model itself), in a fresh context, to judge whether the candidate is correct from $(I_i,q_i,\hat a_i)$. Its decision is
denoted by $\hat y_i\in\{0,1\}$, where $\hat y_i=1$ denotes acceptance. The reference answer is used only to determine $y_i$ and is never provided to the verifier. Verification is therefore evaluated as binary discrimination between correct and incorrect candidates. We report Accuracy, Balanced Accuracy (BAcc.), and $\mathrm{FPR}=\Pr(\hat y=1\mid y=0)$. Directionality is summarized by

\vspace{-1mm}
\[
\mathrm{Bias}=\frac{\mathrm{FP}-\mathrm{FN}}{N},\qquad
\mathrm{D\mbox{-}Skew}=
\frac{(\mathrm{FP}-\mathrm{FN})^2}
{\mathrm{FP}^2+\mathrm{FN}^2+(\mathrm{TP}+\mathrm{TN})(\mathrm{FP}+\mathrm{FN})}.
\]
\vspace{-1mm}

The \verimap{} covers six open-weight VLMs: Qwen3-VL-8B~\citep{bai2025qwen3}, Gemma-4-E4B~\citep{team2026gemma}, Phi-4-Multimodal~\citep{abouelenin2025phi}, MedGemma-1.5-4B~\citep{sellergren2025medgemma}, Lingshu-7B~\citep{xu2026lingshu}, and HuatuoGPT-Vision-7B~\citep{chen2024towards}. We evaluate them on five datasets: VQA-RAD~\citep{lau2018dataset}, SLAKE~\citep{liu2021slake}, PathVQA~\citep{he2020pathvqa}, MedXpert-QA~\citep{zuo2025medxpertqa}, and PMC-VQA~\citep{zhang2023pmc}. The details are provided in Appendix~\ref{app:experimental_setup}.


\subsection{The Verification Mirage}
\label{sec:verification_mirage}

\begin{figure}[htbp]
    \centering
    \includegraphics[width=\linewidth]{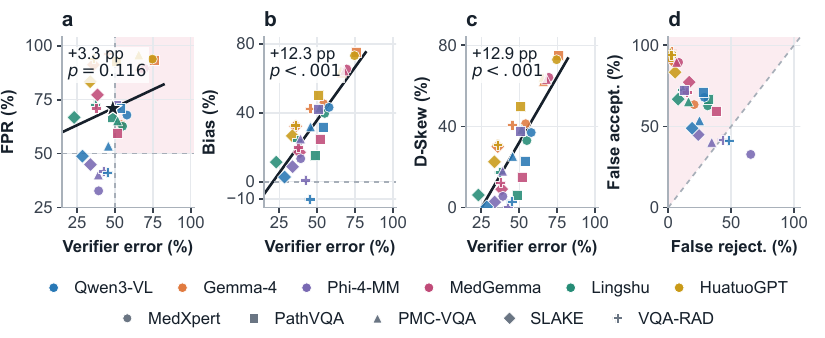}
    \caption{\textbf{Analysis of the Verification Mirage across model-dataset settings.} \textbf{(a-c)} Verifier error versus FPR, Bias, and D-Skew across the 30 settings. \textbf{(d)} False acceptance versus false rejection. Solid lines in \textbf{(a-c)} show fitted trends.}
    \label{fig:verification_mirage}
\end{figure}

\paragraph{Self-verification frequently accepts incorrect answers.}
Fig.~\ref{fig:verification_mirage} shows that verification is both inaccurate and  permissive across 6 models and 5 datasets (30 settings).
The star in panel (a) marks the macro average over these settings: verifiers are wrong about half the time ($48.5\%$ error), and even among incorrect candidates they return \textsc{Correct} $71.0\%$ of the time.
Across settings, verifier error tends to be positively correlated with acceptance-side failure in Bias and D-Skew. To quantify these trends while accounting for systematic differences across models and datasets, we fit separate linear mixed-effects models with verifier error as the fixed-effect predictor and random intercepts for model and dataset.
The fitted regression coefficients of the fixed effect in panels (a)-(c) show that each 10 percentage-point increase in verifier error is associated with increases of $3.3$, $12.3$, and $12.9$ points in FPR, Bias, and D-Skew, respectively (the FPR slope is has $p=.116$, while both asymmetry slopes have $p<.001$  with details in Appendix~\ref{app:glmm}).
Panel (d) makes the directional asymmetry explicit: FPR exceeds FNR in 27 of 30 settings.
Verification is therefore not only inaccurate, but its errors are also usually skewed toward accepting incorrect candidates.
We call this combination of substantial verifier error and directional false acceptance the \emph{Verification Mirage}.

\subsection{Generator-Verifier Error Coupling}
\label{sec:error_inheritance}

\paragraph{Models are more likely to accept their own wrong answers.}

The acceptance asymmetry in Fig.~\ref{fig:verification_mirage} could arise from a generally permissive verifier, without any dependence on who produced the error. We therefore ask a stricter question: \emph{among candidates already known to be wrong, does a verifier accept its own errors more often than errors from other models?} To isolate this dependence, every verifier judges frozen wrong candidates from all six generators across five datasets, under the same prompt and in a fresh context, with generator identity hidden.

\begin{figure}[htbp]
    \centering
    \includegraphics[width=0.65\linewidth]{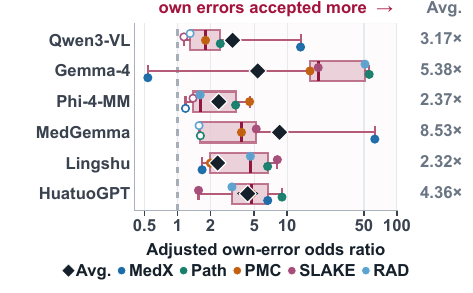}
    \captionsetup{width=0.75\linewidth}
    \caption{\textbf{Own-error dependence across verifiers and datasets.} Points show dataset-specific adjusted odds ratios, and black diamonds show five-dataset pooled estimates with 95\% CIs.}
    \label{fig:error_inheritance}
\end{figure}

We fit a mixed-effects model that separates this own-error effect from two
simpler sources of variation: some verifiers may be more permissive overall, and some generators may produce errors that are easier to accept regardless of who verifies them. The resulting self-effect odds ratio therefore measures the
additional odds of accepting a wrong candidate when generator and verifier are
the same model. An odds ratio above one indicates preferential acceptance of
the verifier's own errors. Appendix~\ref{app:glmm} gives the full specification.
All six verifiers show this own-error dependence, with adjusted ORs from $2.32$ to $8.53$ and all Holm-adjusted $p<.001$ in Fig.~\ref{fig:error_inheritance}. In other words, after accounting for general verifier permissiveness and differences among generators' errors, the odds of accepting a wrong answer are consistently higher when that answer was produced by the verifier itself.

\subsection{From Existing Explanations to a Discriminative Check}
\label{sec:simpler_explanations}

\begin{figure}[htbp]
    \centering
    \includegraphics[width=\linewidth]{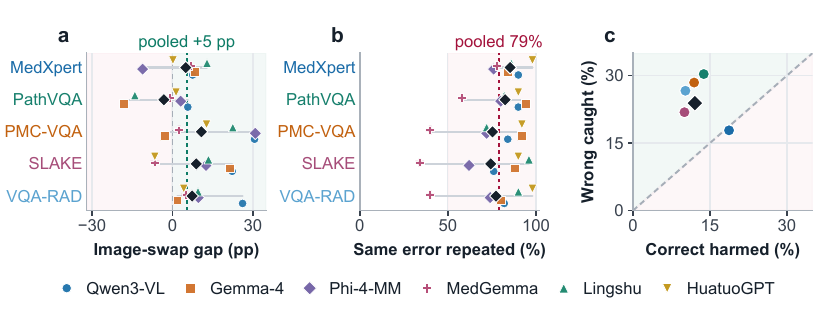}
    \caption{\textbf{Analysis of existing explanations and a discriminative visual check.} \textbf{(a)} Image-swap gap. \textbf{(b)} Same error repeated under candidate-hidden re-solving. \textbf{(c)} Wrong caught versus correct harmed under the falsifiable visual-check probe.}
    \label{fig:existing_explanations}
\end{figure}

Prior work suggests two potential explanations for the \emph{Verification Mirage}. On the \emph{evidence side}, insufficient visual grounding can produce hallucinated judgments~\citep{lee2024volcano,liao2025can}. On the \emph{decision side}, conditioning on a proposed answer can induce agreement bias and over-validation~\citep{andrade2025let}. We probe these explanations with two interventions. Image replacement tests whether verifier decisions are selectively tied to the current image, while candidate hiding tests whether removing direct access to the proposed answer breaks the original error.

\paragraph{Using the correct image does not reliably separate correct from incorrect answers.}
We replace the original image with its nearest non-identical neighbor in BioMedCLIP image-embedding space, restricting the search to the same dataset
and task stratum while holding all other inputs fixed. Across the full model-dataset matrix, $66.4\%$ of false acceptances persist after replacement,
compared with $70.4\%$ of true acceptances (Fig.~\ref{fig:existing_explanations}a). Thus, changing the image
can alter the verdict without preferentially removing incorrect acceptances. PMC-VQA makes this distinction especially clear: image replacement preserves
$35.3\%$ of false acceptances and $38.3\%$ of true acceptances.

\paragraph{Re-solving without the candidate often repeats the same error.}
Among known-wrong candidates, candidate-hidden re-solving produces an answer judged identical to or compatible with the original error in $79.0\%$ of
cases (Fig.~\ref{fig:existing_explanations}b). Removing direct access to the candidate therefore reduces one source of anchoring, but does not reliably produce a competing hypothesis that can falsify the original answer.

Together, the two interventions isolate a narrower failure mode. Verification can depend on the image and can be made candidate-independent, yet still fail to discriminate correct from incorrect acceptances. What is missing is a candidate-specific visual observation that can falsify the answer and govern
the verdict.

\paragraph{A falsifiable visual check recovers selective verification.}
The preceding results suggest why self-verification fails to obtain the usual advantage of checking over solving. Seeing the image again or removing the
candidate still leaves the model with much of the original visual problem. We therefore test a narrower requirement: \emph{can the candidate be converted into a small visual distinction whose observed outcome directly determines whether
the candidate should remain accepted?}

We evaluate this requirement with a frozen mechanism probe. When an accepted candidate disagrees with a candidate-hidden hypothesis, the pair is used only
to specify the smallest visual distinction between them. A fresh observer then measures that distinction without seeing either hypothesis or its role, and a fixed mapping determines whether the sealed observation supports the candidate. The alternative therefore defines what must be tested, but does not act as a vote.

The diagnostic signature is selective revocation: incorrect acceptances should be caught more often than correct acceptances are harmed. In Fig.~\ref{fig:existing_explanations}c, this holds in 25 of 30 model–dataset settings. Pooled across cells, the probe catches $23.9\%$ of false acceptances while harming $12.1\%$ of true acceptances, corresponding to 2,809 rescues and 1,064 harms. The probe therefore provides evidence consistent with a candidate mechanism for selective verification: reducing the candidate to a falsifiable visual test and tying the verdict to the resulting observation.

\paragraph{False acceptance can carry errors across self-improvement rounds.}
When verification is reused as feedback in a five-turn self-improving actor-verifier loop, a local false acceptance can become a persistent positive
signal that prevents correction. As shown in Fig.~\ref{fig:agent_repair}, all five datasets, $66.9$-$96.0\%$ of initially wrong answers finish both wrong and falsely accepted, while only $1.5$-$5.9\%$ are corrected. Cumulative false acceptance further increases
by $9.1$-$15.2$ percentage points on average across datasets. These results expose a system-level risk for verifier-driven self-improvement: repeated feedback does not necessarily wash out verification errors and can instead lock them into subsequent rounds. Appendix~\ref{app:iterative_reuse} provides
the turn-by-turn stress test. Section~\ref{sec:agent_repair} evaluates whether repairing the verification channel weakens this lock-in without sacrificing useful self-correction.

%% file: sections/experiments.tex
\section{\ours{}: Intervening and Repairing Self-Verification}
\label{sec:experiments}

\subsection{\ours{}}
\label{sec:miragecheck}
\begin{figure}[t]
    \centering
    \includegraphics[width=\linewidth]{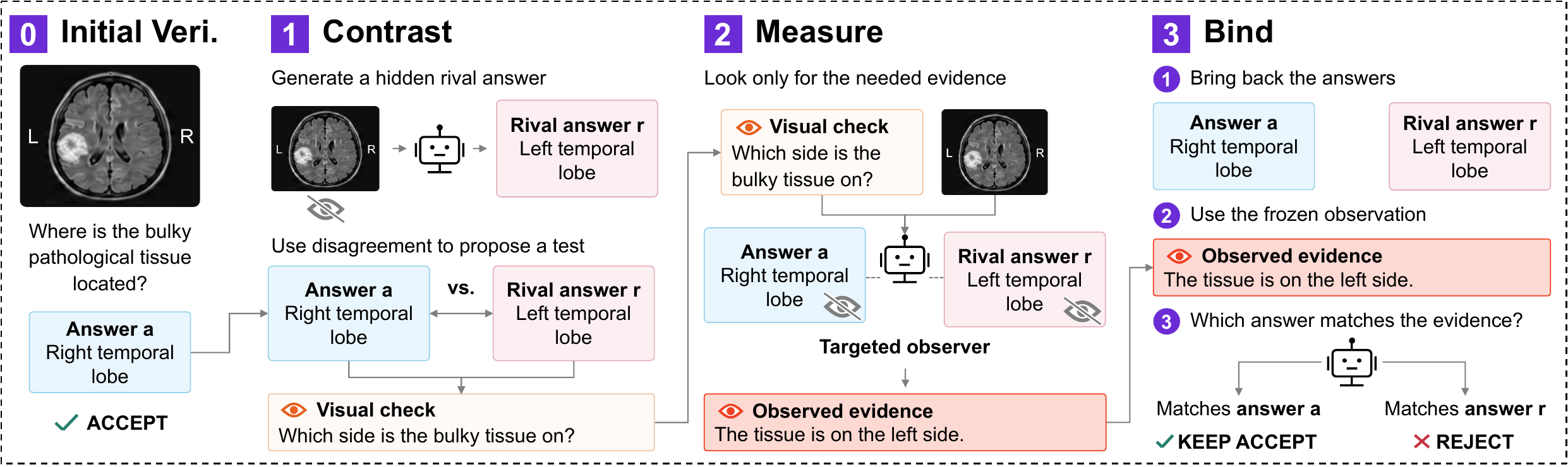}
    \caption{\textbf{Overview of \ours{}.} \ours{} rechecks initially accepted candidates through \textsc{Contrast}, \textsc{Measure}, and \textsc{Bind}, which define a visual distinction, observe candidate-hidden evidence, and bind that evidence back to the competing hypotheses.}
    \label{fig:method}
\end{figure}

Section~\ref{sec:simpler_explanations} motivates a more specific design principle for addressing the Verification Mirage. Verification needs an answer-discriminative visual test, not another solution of the same VQA. Candidate-independent re-solving can still reproduce the original error. Disagreement becomes useful when it specifies \emph{what the image must distinguish} and the resulting evidence governs the verdict. \ours{} operationalizes this principle by separating test construction, visual measurement, and evidence-to-verdict binding (Fig.~\ref{fig:method}). It is applied to an initially accepted candidate $a$ for efficiency. In \textsc{Contrast}, the VLM first re-answers the image and question in a fresh context without seeing $a$, producing a rival $r$. Their answer-critical disagreement is then converted into a neutral visual check that distinguishes
the two hypotheses. In \textsc{Measure}, a fresh observer receives only the image and this check, with both answers hidden, and records the relevant visual evidence. This observation is frozen before either hypothesis is revealed.
In \textsc{Bind}, the frozen evidence is paired with $a$ and $r$ in randomized order to determine which answer it supports. The original acceptance is kept only when the evidence supports $a$. Otherwise, it is revoked.

This information separation prevents the rival from acting as a second vote. It can define what should be measured, but the verdict is determined only after the evidence has been formed independently of both answers. Initial rejections
are unchanged, making \ours{} a targeted intervention on the false-acceptance failure identified by \verimap{}. All stages use the same frozen VLM checkpoint in fresh contexts and require no additional training, reference answer, or external critic at inference time. Appendix~\ref{app:miragecheck_spec} gives the full details about it.

\subsection{Evaluation Protocol and Main Results}

\paragraph{Evaluation setup and baselines.} We evaluate all methods on the same frozen candidates produced by the six VLMs across the five medical VQA datasets in \verimap{}. For each candidate, the image, question, model checkpoint, and proposed answer are fixed. Ground-truth answers are used only for post-hoc evaluation. We retain the four metrics from Section~\ref{sec:problem_formulation}. This paired setup isolates changes in the verification procedure without changing the underlying generations. For method comparison, we use BAcc. as the primary performance metric. A method can reduce FPR simply by rejecting more candidates, at the cost of increasing false rejection. Because BAcc. weights correct-candidate preservation and incorrect-candidate rejection equally, it measures whether the reduction in false acceptance yields a favorable overall trade-off.

The baselines target the two categories motivated by Section~\ref{sec:simpler_explanations}. For candidate independence, \emph{Candidate-Hidden Re-solving} hides the candidate and re-solves the original image-question pair in a fresh context before comparison~\citep{manakul2023selfcheckgpt}, while \emph{SGV} forms candidate-independent visual findings before judging the revealed candidate~\citep{andrade2025let}. For stronger visual grounding, we adapt two training-free hallucination mitigation methods to verification: \emph{CausalLens} strengthens visually sensitive attention pathways during decoding~\citep{ji2026causallens}, while \emph{ClearSight} amplifies visual signals during multimodal fusion~\citep{yin2025clearsight}. Vanilla self-verification provides the baseline reference. Together, these baselines test whether candidate independence or stronger visual grounding is sufficient without the evidence-bound information flow used by \ours{}. Implementation details are provided in Appendix~\ref{app:experimental_setup}.

\definecolor{resultblue}{HTML}{A8D0F7}

\begin{table}[!t]
\centering
\caption{\textbf{Verification performance across five medical VQA datasets.}
Entries are unweighted six-model means (\%); BAcc is $(\mathrm{TPR}+\mathrm{TNR})/2$. Arrows indicate the preferred direction. $^{\dagger}$Candidate-independent and $^{\ddagger}$visual-grounding baselines.}
\label{tab:miragecheck_main}
\vspace{-0.2\baselineskip}
\setlength{\tabcolsep}{2.6pt}
\renewcommand{\arraystretch}{0.97}
\scriptsize
\begin{tabular}{@{}lcccccccccc@{}}
\toprule
\multirow{2}{*}{Method} & \multicolumn{2}{c}{\textbf{PathVQA}} & \multicolumn{2}{c}{\textbf{VQA-RAD}} & \multicolumn{2}{c}{\textbf{SLAKE}} & \multicolumn{2}{c}{\textbf{MedXpert-QA}} & \multicolumn{2}{c}{\textbf{PMC-VQA}} \\
\cmidrule(lr){2-3} \cmidrule(lr){4-5} \cmidrule(lr){6-7} \cmidrule(lr){8-9} \cmidrule(lr){10-11}
 & BAcc\,$\uparrow$ & FPR\,$\downarrow$ & BAcc\,$\uparrow$ & FPR\,$\downarrow$ & BAcc\,$\uparrow$ & FPR\,$\downarrow$ & BAcc\,$\uparrow$ & FPR\,$\downarrow$ & BAcc\,$\uparrow$ & FPR\,$\downarrow$ \\
\midrule
Vanilla verification & 53.45 & 73.77 & 55.39 & 73.55 & 59.72 & 68.18 & 52.00 & 76.13 & 57.04 & 72.92 \\
\cdashline{1-11}[0.35pt/2pt]\addlinespace[1.5pt]
Candidate-Hidden Re-solving$^{\dagger}$ & 54.75 & 83.36 & 52.38 & 92.58 & 51.70 & 94.75 & 49.99 & 91.69 & 54.08 & 85.64 \\
SGV$^{\dagger}$ & 56.14 & 58.05 & 58.98 & 62.12 & 60.40 & 61.57 & \textbf{52.48} & 56.38 & 60.92 & 56.35 \\
\cdashline{1-11}[0.35pt/2pt]\addlinespace[1.5pt]
CausalLens$^{\ddagger}$ & 52.54 & 67.59 & 56.07 & 66.59 & 57.91 & 60.65 & 51.94 & 59.72 & 56.68 & 65.47 \\
ClearSight$^{\ddagger}$ & 52.82 & 65.26 & 55.30 & 64.38 & 58.18 & 58.33 & 52.06 & 61.94 & 55.93 & 65.88 \\
\cdashline{1-11}[0.35pt/2pt]\addlinespace[1.5pt]
\rowcolor{resultblue!20}\ours{} (ours) & \textbf{62.07} & \textbf{32.40} & \textbf{63.65} & \textbf{42.26} & \textbf{63.47} & \textbf{44.75} & 51.75 & \textbf{42.17} & \textbf{65.27} & \textbf{35.97} \\
\midrule
 & $|$Bias$|$\,$\downarrow$ & D-Skew\,$\downarrow$ & $|$Bias$|$\,$\downarrow$ & D-Skew\,$\downarrow$ & $|$Bias$|$\,$\downarrow$ & D-Skew\,$\downarrow$ & $|$Bias$|$\,$\downarrow$ & D-Skew\,$\downarrow$ & $|$Bias$|$\,$\downarrow$ & D-Skew\,$\downarrow$ \\
\midrule
Vanilla verification & 38.71 & 32.19 & 22.99 & 17.03 & 16.45 & 10.67 & 54.23 & 50.70 & 49.56 & 46.50 \\
\cdashline{1-11}[0.35pt/2pt]\addlinespace[1.5pt]
Candidate-Hidden Re-solving$^{\dagger}$ & 48.91 & 46.69 & 38.32 & 36.92 & 32.50 & 31.30 & 68.88 & 67.69 & 60.58 & 59.36 \\
SGV$^{\dagger}$ & 26.94 & 20.86 & 19.66 & 15.39 & 13.71 & 9.56 & 34.71 & 28.72 & 35.11 & 30.40 \\
\cdashline{1-11}[0.35pt/2pt]\addlinespace[1.5pt]
CausalLens$^{\ddagger}$ & 31.50 & 24.04 & 19.70 & 13.32 & 14.42 & 8.81 & 37.74 & 33.35 & 42.00 & 37.95 \\
ClearSight$^{\ddagger}$ & 34.63 & 27.51 & 17.89 & 12.40 & 20.15 & 13.86 & 40.23 & 36.48 & 41.93 & 38.56 \\
\cdashline{1-11}[0.35pt/2pt]\addlinespace[1.5pt]
\rowcolor{resultblue!20}\ours{} (ours) & \textbf{12.99} & \textbf{7.08} & \textbf{11.38} & \textbf{4.68} & \textbf{8.00} & \textbf{3.31} & \textbf{20.35} & \textbf{16.27} & \textbf{19.26} & \textbf{13.94} \\
\bottomrule
\end{tabular}
\end{table}

\paragraph{\ours{} improves verification beyond existing baselines.}
Table~\ref{tab:miragecheck_main} compares \ours{} with Vanilla verification, candidate-independent baselines, and visual-grounding baselines across all five
datasets. \ours{} yields the lowest FPR, $|\mathrm{Bias}|$, and D-Skew on every
dataset, while achieving the highest balanced accuracy on four of five. Compared with
SGV, the strongest competing baseline by mean balanced accuracy, \ours{} improves
the 30-setting mean from $57.78\%$ to $61.24\%$ and reduces mean FPR from
$58.89\%$ to $39.51\%$. Balanced accuracy improves over SGV on PathVQA,
VQA-RAD, SLAKE, and PMC-VQA, but decreases on MedXpert-QA from $52.48\%$ to
$51.75\%$, which is also slightly below Vanilla, CausalLens, and ClearSight
($51.94$--$52.06\%$). On that dataset, FPR still falls from $56.38\%$ to $42.17\%$.

\subsection{Mechanistic Analysis and Ablations}
\paragraph{\ours{} reduces false acceptance while improving balanced discrimination.}

\begin{figure}[htbp]
    \centering
    \includegraphics[width=\linewidth]{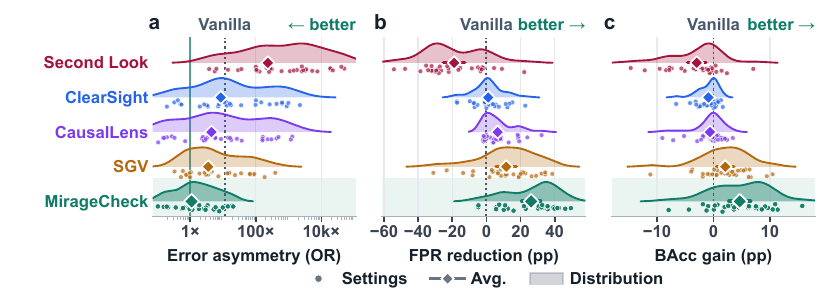}
    \caption{\textbf{Comparison of verification methods across 30 model-dataset settings.} \textbf{(a)} False-acceptance asymmetry, measured by $\operatorname{odds}(\mathrm{FPR})/\operatorname{odds}(\mathrm{FNR})$. Diamonds show adjusted pooled estimates with 95\% confidence intervals. \textbf{(b)} FPR reduction relative to Vanilla. \textbf{(c)} BAcc gain relative to Vanilla. Points show individual settings, curves show their distributions, and diamonds in \textbf{(b,c)} show means with $\pm1.96$ standard errors.}
    \label{fig:miragecheck_boundary}
    \vspace{-2mm}
\end{figure}

Table~\ref{tab:miragecheck_main} establishes the aggregate improvement. We next
ask whether \ours{} changes the specific failure isolated in Section~\ref{sec:simpler_explanations}, rather than simply shifting the overall
acceptance rate. Fig.~\ref{fig:miragecheck_boundary} examines this question across the 30 model and dataset settings. Under Vanilla verification, errors remain strongly skewed toward false acceptance. Candidate-Hidden Re-solving and the visual-grounding interventions leave substantial directional error, while SGV reduces but does not eliminate it. \ours{} reduces false acceptance further.

As presented in Fig.~\ref{fig:miragecheck_boundary}, this reduction is accompanied by higher balanced accuracy. Among all settings, \ours{} lowers FPR by $14.63$ points and raises BAcc by $2.57$ points relative to SGV. Relative to Vanilla, both metrics also improve, as shown in Fig.~\ref{fig:miragecheck_boundary}. The joint reduction in false acceptance and increase in BAcc is important because a uniformly more conservative verifier could lower FPR simply by rejecting more candidates. Instead, \ours{} substantially reduces the acceptance-side asymmetry while preserving useful acceptances across most settings.

\paragraph{Removing any \ours{} stage weakens performance.}

We ablate one stage of \ours{} at a time on a PathVQA subset while keeping the remaining procedure fixed. Each variant removes only the designated stage and introduces no substitute inference.

\begin{wraptable}{r}{0.61\linewidth}
\centering
\caption{\textbf{Ablation of \ours{}.}}
\label{tab:miragecheck_ablation}
\vspace{-0.3\baselineskip}
\setlength{\tabcolsep}{3pt}
\renewcommand{\arraystretch}{1.02}
\footnotesize
\begin{tabular}{@{}lrrrr@{}}
\toprule
Variant & BAcc $\uparrow$ & FPR $\downarrow$ &
$|\mathrm{Bias}| \downarrow$ & D-Skew $\downarrow$ \\
\midrule
Full \ours{}              & \textbf{58.28} & \textbf{52.98} & \textbf{20.08} & \textbf{14.72} \\
w/o \textsc{Contrast}     & 53.49 & 72.42 & 35.27 & 29.67 \\
w/o \textsc{Measure}      & 54.61 & 68.80 & 32.60 & 26.46 \\
w/o \textsc{Bind}         & 45.38 & 68.80 & 25.15 & 17.63 \\
\bottomrule
\end{tabular}
\end{wraptable}

Table~\ref{tab:miragecheck_ablation} shows that the full pipeline performs best on all four metrics. Removing \textsc{Contrast} produces the largest drop, reducing BAcc. from $58.28\%$ to $53.49\%$ and increasing FPR from
$52.98\%$ to $72.42\%$. Removing \textsc{Measure} or replacing \textsc{Bind} with the baseline also degrades performance, reducing BAcc to $54.61\%$ and
increasing FPR to $68.80\%$. The same pattern holds for $|\mathrm{Bias}|$ and D-Skew. These results suggest that the gain is not obtained from contrast construction alone. It depends on carrying the contrast through visual measurement and into the final decision.

\subsection{\ours{} in the self-improving Actor-Verifier Loop}
\label{sec:agent_repair}

\paragraph{\ours{} reduces false feedback during iterative self-correction.}

\begin{figure}[htbp]
    \centering
    \includegraphics[width=\linewidth]{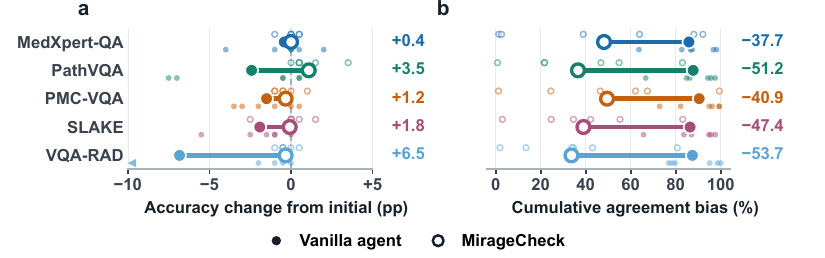}
    \caption{\textbf{\ours{} in the self-improving actor-verifier loop.} Matched comparison of Vanilla verification and \ours{} in the same five-turn feedback loop, averaged over six models within each dataset. Filled and open markers denote Vanilla and \ours{}, respectively. \textbf{(a)} Final answer accuracy relative to the initial answer. \textbf{(b)} Cumulative false acceptance among initially incorrect trajectories. Labels show the paired \ours{} minus Vanilla difference in percentage points.}
    \label{fig:agent_repair}
\end{figure}

Section~\ref{sec:simpler_explanations} showed that false acceptance can become a persistent feedback signal when verification is reused for self-improvement. We therefore rerun the same five-turn self-improving actor-verifier loop and replace only Vanilla verification with \ours{}. Fig.~\ref{fig:agent_repair} reports the matched comparison across all six models and five datasets. Panel (a) measures final accuracy relative to the initial answer, while panel (b) measures cumulative false acceptance among trajectories that were initially incorrect.

With Vanilla feedback, average accuracy falls from $41.20\%$ initially to $38.58\%$ after five turns. With \ours{}, it instead remains at $41.27\%$. The same pattern holds at the dataset-average level, with \ours{} achieving higher final accuracy than Vanilla on all five datasets in Fig.~\ref{fig:agent_repair}a. More importantly, cumulative false acceptance falls from $87.67\%$ to $42.72\%$, with reductions of $37.7$ to $53.7$ percentage points across datasets (panel b). These results highlight a key limitation of vanilla verifier-driven self-improvement. Repeated feedback is useful only when the verification signal itself is reliable. \ours{} does not produce large gains through repeated correction. Instead, it prevents false verification signals from locking initial errors into subsequent self-improvement rounds.

%% file: sections/conclusion.tex
\section{Conclusion}

We study when visual self-verification provides a discriminative check rather than another judgment of the same problem. \verimap{} reveals a \emph{Verification Mirage}: medical VLMs frequently accept incorrect candidates, with substantial generator--verifier error coupling when the same model fills both roles. Our analyses show that candidate independence and stronger visual grounding alone are insufficient, and support a design principle in which verification tests an answer-discriminative visual distinction and ties the verdict to the resulting evidence. \ours{} operationalizes this principle through \textsc{Contrast}, \textsc{Measure}, and \textsc{Bind}, substantially reducing false acceptance while improving balanced accuracy on average across the evaluation suite. Overall, reliable visual self-verification requires making a candidate falsifiable from the current image rather than simply asking the model to judge again.

%% file: sections/statements.tex
\subsection*{AI use statement}
Language models assisted with drafting and figure prototyping. Authors inspected the outputs and retain responsibility for the design, analysis, interpretation, and claims.

\subsection*{Reproducibility statement}
To support reproducibility, we provide detailed experimental specifications
throughout the paper and appendix. Appendix~B reports the evaluation suite,
exact model checkpoints, metrics and coverage handling, correctness
label validation, and inference settings. Appendices~C and~D provide the complete
statistical analyses and controlled diagnostic experiments. Appendix~E specifies
the full \ours{} protocol and its information-access contract. Appendices~F
and~G report supplementary evaluations, controls, ablations, and the iterative
actor-verifier experiments. Appendix~H provides all prompts, parsing rules, and
reporting procedures. All experiments use open-source models and fixed evaluation procedures to support matched comparisons. We will publicly release the implementation and evaluation code upon publication.

%% file: sections/appendix.tex
\numberwithin{figure}{section}
\numberwithin{table}{section}

\noindent\textbf{Appendix roadmap.}
Appendix~\ref{app:related_work} provides extended related work and positions our study relative to prior work on self-verification, model-based feedback, multimodal verification, and medical VLM reliability. Appendix~\ref{app:experimental_setup} describes the evaluation suite, motivates the focus on medical VQA, defines the evaluation metrics, and provides implementation and correctness-labeling details. Appendix~\ref{app:verimap_full} reports additional \verimap{} analyses, including task-level results and controlled tests of preferential acceptance of a model's own errors. Appendix~\ref{app:mechanism_diagnostics} tests whether the observed failure can be explained by visual access, candidate conditioning, or additional visual information, and examines the role of discriminative evidence. Appendix~\ref{app:miragecheck_spec} describes the \ours{} protocol and its information-separation design. Appendix~\ref{app:miragecheck_results} provides additional \ours{} results and controls, including results across models and datasets, matched-rejection controls, and information-hiding ablations. Appendix~\ref{app:iterative_reuse} studies how verification errors behave when verifier feedback is reused across multiple turns. Finally, Appendix~\ref{app:prompts_implementation} provides the prompts, semantic-judge instructions, limitations, and broader impact.

\section{Extended Related Work}
\label{app:related_work}
\paragraph{Self-verification, self-correction, and verifier reliability.}
Self-verification and self-evaluation have been used to filter, select, or revise model outputs across reasoning, code, and multimodal tasks~\citep{kadavath2022language,ren2023self,ni2023lever,chen2023teaching,sun2025mm,wu2025generate}. However, additional model calls do not necessarily provide a reliable corrective signal: intrinsic self-correction can fail without informative external feedback~\citep{huang2024large}, while repeated sampling or self-checking may reproduce rather than resolve an original error~\citep{manakul2023selfcheckgpt}. Factored Chain-of-Verification (CoVe) similarly plans fact-checking questions from an initial response and answers them in independent contexts before revising the response \citep{dhuliawala2024chain}. While it shares the goal of reducing dependence on the original answer, CoVe targets text-only factual correction and does not require verification questions to isolate an answer-discriminative visual distinction or constrain a frozen-candidate verdict by sealed visual
evidence. Our diagnostics show that candidate independence alone is insufficient
in medical VQA. Model-based judges can also exhibit agreement and source-dependent biases, including preference for their own generations
\citep{xu2024pride,panickssery2024llm,wataoka2024self,andrade2025let}. Our focus is distinct from asking whether another inference pass improves final
task accuracy or whether a judge prefers its own outputs in aggregate. \verimap{} conditions explicitly on candidate correctness and tests whether a verifier is especially likely to accept \emph{its own wrong answers and investigate why}. Its complete cross-generator design separates this own-source excess from general verifier permissiveness and generator-specific error acceptability, while
a correctness-balanced source analysis further probes whether the effect is specific to errors rather than general own-source preference. We therefore treat
known self-preference as a competing explanation for generator-verifier dependence, rather than equating the two.

\paragraph{Verifier feedback and self-improving systems.}
Verifier outputs increasingly serve not only as post-hoc judgments but also as
feedback for subsequent generation or optimization
\citep{gou2024critic,yuan2024self,liu2025agent0}.
Self-rewarding methods use model-generated judgments as optimization signals,
while self-evolving vision-language agents can explicitly couple solver,
verifier, and repair roles. This reuse changes the consequence of verification
error: a false acceptance is no longer an isolated evaluation mistake, but can
become an endogenous positive signal that influences later behavior.
Prior work establishes both the promise and the limitations of model-generated
feedback, whereas our analysis isolates a complementary reliability condition:
whether verification supplies information that can discriminate an actor's error
rather than reproduce it. Our multi-turn experiments are therefore a bounded
consequence analysis of the failure identified by \verimap{}, not a claim that
all verifier-driven self-improvement degrades. More broadly, the results motivate
a reliability requirement for self-improving systems: verification should be
evaluated not only by average judgment accuracy, but by whether its errors remain
coupled to the failures of the actor whose behavior it supervises.

\paragraph{Multimodal verification and discriminative visual evidence.}
Multimodal reliability methods improve evaluation or correction through
candidate-independent reasoning, stronger visual grounding, retrospective
verification, or learned visual reward models
\citep{lee2024volcano,andrade2025let,sun2025mm,
wu2025generate,wang2025visualprm,ji2026causallens,yin2025clearsight}. Woodpecker similarly formulates questions and validates visual facts before
correcting an MLLM response~\citep{yin2024woodpecker}. However, it is a post-hoc correction pipeline that uses separate pretrained components to construct visual
evidence, whereas our controlled setting asks whether the same frozen VLM can
verify its own candidate without an external critic. Moreover, \ours{} uses candidate-rival disagreement only to specify a falsifiable distinction, hides both hypotheses during measurement, and binds the frozen evidence back to
anonymous hypotheses before deciding the verdict.
Most closely, Self-Grounded Verification reduces candidate-induced agreement by eliciting candidate-independent information before judging the revealed answer~\citep{andrade2025let}.
These approaches motivate two natural remedies for visual verification failure: candidate independence and stronger visual grounding. \verimap{} separates these from a stricter requirement: a verifier can
reason independently of the candidate and use the correct image while still
failing to obtain evidence that distinguishes the candidate from its alternatives.
\ours{} operationalizes this diagnosis in the same-model, frozen-candidate
setting without additional training or an external critic. A candidate-rival
contrast specifies what must be tested, the corresponding visual observation is
formed with both answers hidden, and the frozen evidence is subsequently bound
back to anonymous hypotheses. This information separation distinguishes
\ours{} from a second vote, another holistic re-solution, or a rubric that remains
conditioned on the competing answers during evidence formation.

\paragraph{Medical VLM reliability and study scope.}
Medical and multimodal reliability work has studied hallucination detection,
uncertainty estimation, visual grounding, retrieval, external clinical evidence,
and specialized verifier or reward models
\citep{mahmood2025phrase,chung2025verifact,liu2024retrieval,
liao2025univrse,wang2025visualprm}.
Such approaches can introduce information unavailable to the original model and
are complementary to our objective. We deliberately study the stricter
self-contained setting in which the same VLM must verify a frozen answer from
the current image and question, without a reference answer, retrieved record,
separately trained verifier, or external clinical critic. This isolates whether
the verification process itself produces a discriminative signal, rather than
whether reliability can be recovered by adding a stronger source of information.
We focus on medical VQA not because we assume the failure is unique to medicine,
but because it provides a revealing testbed for this reliability boundary.
As motivated in the main text, many medical VQA questions hinge on fine-grained
but auditable distinctions in findings, anatomy, laterality, counts, or spatial
relations, allowing the evidence required to test a candidate to be manipulated
and inspected explicitly. Our sampled cross-domain comparison additionally
places the medical benchmarks in a substantially higher false-acceptance regime
than the sampled general-domain benchmarks. This combination makes medical VQA
useful both for exposing the failure and for localizing what evidence changes
verification. We treat the cross-domain result as motivation rather than evidence
of a uniquely medical effect; Appendix~\ref{app:medical_testbed} provides the
corresponding analysis and scope limitations.

\section{Experimental Setup and Evaluation Protocol}
\label{app:experimental_setup}
\begin{table}[htbp]
\centering
\small
\caption{\textbf{Open-weight VLM checkpoints used in the unified audit.}
We report the paper-facing model name together with the exact checkpoint identifier.}
\label{tab:model_suite}
\begin{tabular}{ll}
\toprule
Model & Checkpoint identifier \\
\midrule
Qwen3-VL-8B-Instruct & \texttt{Qwen/Qwen3-VL-8B-Instruct} \\
Gemma-4-E4B-it & \texttt{google/gemma-4-E4B-it} \\
Phi-4-multimodal-instruct & \texttt{microsoft/Phi-4-multimodal-instruct} \\
MedGemma-1.5-4B-it & \texttt{google/medgemma-1.5-4b-it} \\
Lingshu-7B & \texttt{lingshu-medical-mllm/Lingshu-7B} \\
HuatuoGPT-Vision-7B-hf & \texttt{FreedomIntelligence/HuatuoGPT-Vision-7B-hf} \\
\bottomrule
\end{tabular}
\end{table}

\subsection{Evaluation Suite}

\begin{figure}[htbp]
    \centering
    \includegraphics[width=0.9\linewidth]{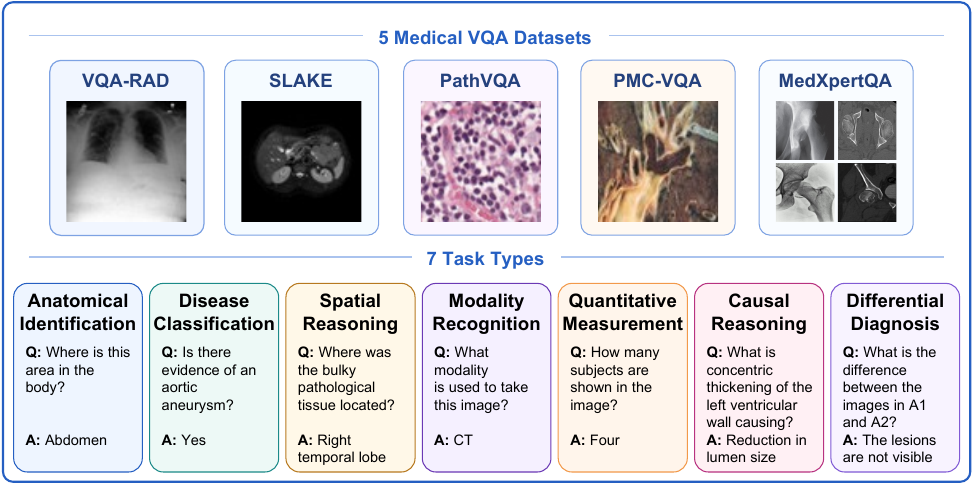}
    \caption{\textbf{Dataset and task coverage of the unified audit.}
    The evaluation spans five medical VQA datasets and seven mutually exclusive
    task strata. Examples illustrate the taxonomy used for stratified analysis;
    the strata characterize task type rather than difficulty.}
    \label{fig:dataset_task_overview}
\end{figure}

The \verimap{} audit spans six open-The audit spans six open-weight VLMs and five medical VQA benchmarks, yielding 30 model-dataset settings. The models are Qwen3-VL-8B~\citep{bai2025qwen3}, Gemma-4-E4B~\citep{team2026gemma}, Phi-4-Multimodal~\citep{abouelenin2025phi}, MedGemma-1.5-4B~\citep{sellergren2025medgemma}, Lingshu-7B~\citep{xu2026lingshu}, and
HuatuoGPT-Vision-7B~\citep{chen2024towards}. The datasets are VQA-RAD~\citep{lau2018dataset}, SLAKE~\citep{liu2021slake}, PathVQA~\citep{he2020pathvqa}, MedXpert-QA~\citep{zuo2025medxpertqa}, and PMC-VQA~\citep{zhang2023pmc}. This $6\times5$ matrix defines the common
evaluation suite used throughout the aggregate analyses. Exact checkpoints are listed in Table~\ref{tab:model_suite}, while
Fig.~\ref{fig:dataset_task_overview} summarizes dataset and task coverage.

\subsection{Why We Focus on Medical VQA}
\label{app:medical_testbed}

\begin{figure}[htbp]
    \centering
    \includegraphics[width=0.94\linewidth]{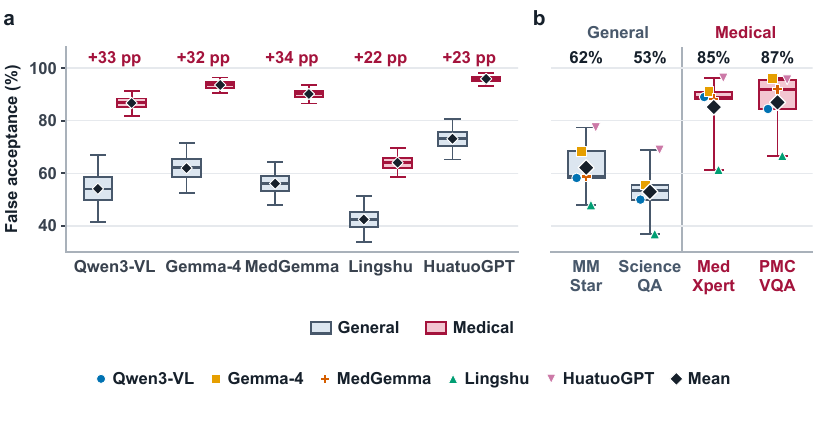}
    \caption{\textbf{False acceptance is higher on the sampled medical VQA benchmarks.} Panel (a) compares false-acceptance rates for five VLMs on sampled general and medical-image benchmarks, with annotations showing the medical minus general difference. Panel (b) shows the corresponding dataset-level distributions across models.}
    \label{fig:medical_testbed_contrast}
\end{figure}

\paragraph{Medical VQA exposes a pronounced false-acceptance regime.}
Figure~\ref{fig:medical_testbed_contrast} provides a descriptive comparison between sampled general- and medical-image benchmarks under self-verification. Across all five displayed VLMs, false acceptance is higher on the medical benchmarks, with medical-general differences ranging from 22 to 34 percentage points. The dataset-level view shows the same separation, with both medical benchmarks occupying a substantially higher false-acceptance regime than the
two general-image benchmarks. It's worth mentioning that this pattern is not induced by adversarial construction: the incorrect candidates are the models' own predictions on the original benchmark image-question pairs. Consistently, in our full medical evaluation, vanilla same-model verification yields dataset-level FPRs of 68.34-75.87\% across the five benchmarks (Table~\ref{tab:miragecheck_main}).

We treat this cross-domain contrast as motivation (rather than evidence of a causal medical-domain effect). The later within-medical diagnostics instead ask what makes these false acceptances difficult to dislodge: candidate-independent re-solving frequently reproduces the original
error, whereas a targeted falsifiable visual check is substantially more selective. Thus, medical VQA provides a setting in which the failure is both pronounced and amenable to controlled analysis of the verification process.

\paragraph{False acceptance is the safety-critical error in medical self-verification.}
False-positive and false-negative errors need not carry equal costs: in decision-theoretic terms, their relative importance depends on the downstream action induced by the prediction. In our verification formulation, the positive
decision is \emph{acceptance}: a false positive therefore endorses an incorrect medical answer as correct, whereas a false negative withholds endorsement from a correct answer. The former can propagate an erroneous finding, anatomical
claim, or diagnosis as a positive correctness signal, and becomes especially consequential when verifier outputs are reused for correction or subsequent
decision making. We therefore treat FPR as the primary directional reliability measure and report BAcc alongside it to assess the tradeoff with false rejection. This symmetric comparison does not assign equal downstream costs to the two errors.

\paragraph{Medical VQA provides auditable answer-critical visual distinctions.}
Beyond exhibiting substantial false acceptance, medical VQA offers a useful setting for isolating what verification must discriminate. The affiliated evaluation in this appendix spans modality recognition, anatomical identification, disease classification, spatial reasoning, quantitative measurement, causal reasoning, and
differential diagnosis (Fig.~\ref{fig:dataset_task_overview}). Across these
tasks, correctness can often be tied to a concrete visual distinction, for example, the presence of a finding, an anatomical location, a count, or a relative spatial relation that can be inspected independently of the
candidate answer. This property enables us to move beyond measuring whether verification fails and ask what visual evidence is sufficient to make the candidate falsifiable. We therefore use medical VQA as a controlled testbed for studying the reliability boundary of visual self-verification.

\subsection{Metrics and Evaluation Coverage}
\label{app:metrics_coverage}

\paragraph{Verification as candidate discrimination.}
For each frozen candidate, the post-hoc label $y\in\{0,1\}$ indicates whether the candidate is correct, while the verifier decision $\hat{y}\in\{0,1\}$ denotes rejection or acceptance. A true positive (TP) is a correct candidate that is accepted, a true negative (TN) is an incorrect candidate that is rejected, a false positive (FP) is an incorrect candidate that is accepted, and a false negative (FN) is a correct candidate that is rejected. Here, ``positive'' refers to verifier acceptance rather than the presence of a clinical
condition.

The binary metrics are
\[
\mathrm{Accuracy}
=
\frac{\mathrm{TP}+\mathrm{TN}}
{\mathrm{TP}+\mathrm{TN}+\mathrm{FP}+\mathrm{FN}},
\qquad
\mathrm{FPR}
=
\frac{\mathrm{FP}}{\mathrm{FP}+\mathrm{TN}},
\qquad
\mathrm{FNR}
=
\frac{\mathrm{FN}}{\mathrm{FN}+\mathrm{TP}}.
\]
For method comparison, our primary performance metric is balanced accuracy:
\[
\mathrm{BAcc}
=\frac{\mathrm{TPR}+\mathrm{TNR}}{2}
=1-\frac{\mathrm{FPR}+\mathrm{FNR}}{2}.
\]
Overall accuracy depends on the proportion of correct and incorrect candidates. BAcc weights correct-candidate preservation and incorrect-candidate rejection equally, so it falls when a reduction in FPR is outweighed by an increase in FNR. An always-reject verifier has BAcc of $50\%$ when both candidate classes are present. 

\paragraph{Directional summaries.}
To characterize whether verification errors are concentrated on false
acceptance, we additionally report
\[
\mathrm{Bias}
=
\frac{\mathrm{FP}-\mathrm{FN}}{N},
\]
where $N=\mathrm{TP}+\mathrm{TN}+\mathrm{FP}+\mathrm{FN}$.
Positive values indicate an excess of false acceptance, negative values
indicate an excess of false rejection, and values near zero indicate little
net directional imbalance.

We also report
\[
\mathrm{D\text{-}Skew}
=
\frac{(\mathrm{FP}-\mathrm{FN})^2}
{\mathrm{FP}^2+\mathrm{FN}^2+
(\mathrm{TP}+\mathrm{TN})(\mathrm{FP}+\mathrm{FN})},
\]
which summarizes the magnitude of directional asymmetry while accounting for
the overall confusion-table composition. Bias and D-Skew are deterministic
summaries of the same confusion table as Accuracy, FPR, and FNR. We therefore
use them as complementary descriptions of error direction rather than as
statistically independent outcomes.

Together, these metrics capture whether verification separates correct from
incorrect candidates, whether its errors are skewed toward false acceptance,
and whether an intervention reduces false acceptance without simply trading
it for false rejection.

\subsection{Validating Answer Correctness}
\label{app:oracle_judge_prompt}

\paragraph{Reference-guided semantic matching.}
Model answers cannot be reliably evaluated by exact string matching. Correct responses may appear as full sentences, abbreviations, synonyms, or equivalent medical phrasing, while exact matching would incorrectly reject such expressions. We therefore assign candidate correctness with a fixed text-only semantic matching procedure after all verifier decisions are frozen. The judge receives only the question, dataset reference answer, and model answer. It returns a binary correctness label under fixed rules that accept clear synonyms and equivalent medical phrasing, apply strict matching to binary questions, and assign no partial credit. The judge never receives the medical image, verifier output, or generator provenance. It therefore serves as a reference-guided answer matcher rather than an inference-time critic. The answer-matching instructions are summarized in Appendix~\ref{app:semantic_judge_prompts}.

\paragraph{Blinded human validation.}
We validate this labeling procedure on a dataset-stratified sample of 100 model predictions, with 20 examples drawn from each of the five datasets.
An expert annotator, blinded to the frozen semantic labels, independently judges whether each model answer is correct relative to the question and
dataset reference. The expert agrees with the semantic judge on 98 of 100 examples, corresponding to 98.0\% agreement and Cohen's $\kappa=0.959$ with a 95\% CI of $[0.898,1.000]$. This audit directly tests whether the automated semantic matching procedure reproduces expert reference-based judgments on a representative sample.

\subsection{Inference and Implementation Details}
All calls within a model-dataset setting use the same checkpoint listed in Table~\ref{tab:model_suite}. Image-conditioned calls receive the current
image, while \textsc{Contrast} and \textsc{Bind} are text-only. Decoding settings are fixed within each method and protocol stage. Temperature-zero decoding is used for deterministic evaluation calls where specified. The first attempt is capped at 512 tokens, with a fixed 1024-token retry when additional output is required for successful parsing. Discrete selection steps use constrained decoding over the allowed choices. For each case, the evaluation runner records the model request, raw response, parsed stage outputs, and protocol variant. Evaluation records are joined to the corresponding frozen source candidate, initial verifier decision, and post-hoc correctness label.

For baseline implementations, Vanilla verification uses the common verifier prompt in Appendix~\ref{app:prompts_implementation}. Candidate-Hidden Re-solving re-answers the image-question pair without the candidate and compares the resulting answer with the frozen candidate. SGV follows the candidate-independent evidence-elicitation procedure of
\citet{andrade2025let} before judging the revealed candidate. CausalLens and ClearSight are adapted from their released training-free visual-grounding procedures and applied during verifier inference with the frozen candidate and common verification prompt. We use the same VLM checkpoint, image, question,
candidate, and decoding settings across matched methods.

\section{Additional \verimap{} Analyses}
\label{app:verimap_full}

\subsection{Results by Task Type}
\label{app:datasets_reasoning}

Medical VQA provides a useful setting for examining the reliability boundary at a finer resolution because its questions span distinct and auditable evidence demands. Broad medical multimodal benchmarks similarly organize evaluation around recurring task families such as anatomical recognition, counting, disease diagnosis, modality understanding, and clinical reasoning~\citep{zuo2025medxpertqa,yip2025medbookvqa}. Guided by these recurring demands and by the task composition of our five benchmarks, we use seven task types that capture the dominant forms of evidence required in our evaluation suite: \emph{modality recognition, anatomical identification, disease classification, spatial reasoning, quantitative measurement, causal reasoning, and differential diagnosis}.

\begin{table}[htbp]
\centering
\small
\caption{\textbf{Task-label agreement across datasets.} Majority consensus is the fraction of questions supported by at least two of the three independent annotators. Annotator agreement is the fraction of individual annotations matching the final consolidated label.}
\label{tab:task_label_agreement}
\begin{tabular}{@{}lrr@{}}
\toprule
Dataset & Majority consensus & Annotator agreement \\
\midrule
VQA-RAD     & 95.8\% & 85.1\% \\
SLAKE       & 97.5\% & 89.9\% \\
PathVQA     & 81.6\% & 73.7\% \\
PMC-VQA     & 94.4\% & 85.6\% \\
MedXpert-QA & 83.7\% & 73.0\% \\
\midrule
All datasets & 91.4\% & 82.1\% \\
\bottomrule
\end{tabular}
\end{table}

The task labels are inferred from the question alone and are independent of generator answers, verifier outputs, and correctness labels. Three independent LLM annotators apply the same predefined taxonomy, followed by a fixed majority voting rule. Across 5,338 questions, 91.4\% receive majority support. We further validate the taxonomy with a blinded human audit. A PhD-level annotator with medical-knowledge training
independently labels 25 randomly sampled questions  and agrees with the consolidated labels on 24 of 25 cases, corresponding to 96.0\% agreement, Cohen's $\kappa=0.953$, and macro-F1 $=0.943$.
Table~\ref{tab:task_label_agreement} reports agreement by dataset. Task stratification is used only to provide a finer diagnostic view of the observed failure. We would like to emphasize that the paper's aggregate verification, directional-error, generator-verifier dependence, and mitigation results do not require task labels. Applying \verimap{} likewise does not require a benchmark to provide a predefined task taxonomy. This part is only provided in the appendix as an additional insight.

\begin{figure}[htbp]
    \centering
    \includegraphics[width=\linewidth]{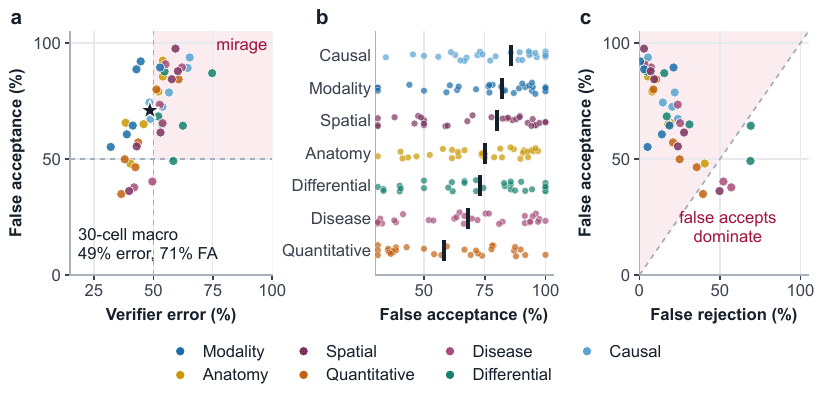}
    \caption{\textbf{Verification reliability across task types.} Panels (a) and (c) show task-model aggregates over five datasets, with the star in panel (a) marking the overall 30-setting macro. Panel (b) shows all 210 task-model-dataset settings, with black ticks marking task medians. Verification behavior varies across task types, but their false-acceptance
    distributions overlap substantially. Across the taxonomy, errors remain predominantly concentrated on false acceptance rather than false rejection. Colors denote task type.}
    \label{fig:task_conditioned_mirage_app}
\end{figure}

Figure~\ref{fig:task_conditioned_mirage_app} shows that task identity adds
resolution to the aggregate reliability picture without changing its central
conclusion. Task-model aggregates occupy different parts of the error and
false-acceptance space, indicating that verification reliability varies with
the evidence demand of the question. At the same time, the 210
task-model-dataset settings overlap substantially, and no task type defines a
uniformly reliable or unreliable regime. The directional pattern is more
consistent: false acceptance remains the dominant error across the taxonomy.
Task stratification therefore refines where the Verification Mirage is most
pronounced, while the underlying acceptance-side failure persists across
reasoning demands.

\paragraph{Task-specific reliability patterns.}
Figure~\ref{fig:task_conditioned_mirage_app} shows how the Verification Mirage varies across the seven reasoning demands. Task-model aggregates occupy different regions of the error and false-acceptance space, indicating that the severity of verification failure depends in part on what evidence the
question requires. This variation is not cleanly separable by task type, however. Across the 210 task-model-dataset settings, false-acceptance
distributions overlap substantially, and no single task defines a uniformly reliable or unreliable regime. The more consistent pattern is directional:
across task types, verification errors remain concentrated on false acceptance rather than false rejection. Task stratification therefore refines where the Verification Mirage is most pronounced, while showing that its acceptance-side failure extends across distinct reasoning demands.

\subsection{False Acceptance vs. False Rejection}
\label{app:glmm}

A high verifier error rate does not reveal how the verifier fails. Our main concern is false acceptance: an incorrect candidate is endorsed as correct. We therefore ask whether verification errors become increasingly skewed toward false acceptance as overall verification becomes less reliable.

We analyze the same 30 model-dataset settings shown in Figure~\ref{fig:verification_mirage}. For each directional metric, we fit a linear mixed-effects model with verifier error as the fixed-effect predictor and random intercepts for model and dataset. Models are fit by REML, and each slope is compared with an intercept-only model using a Kenward-Roger test.

For every 10 percentage-point increase in verifier error, FPR increases by 3.3 points ($p=.116$), signed Bias increases by 12.3 points ($p<.001$), and D-Skew increases by 12.9 points ($p<.001$). The FPR slope is not statistically significant under the Kenward-Roger test, while both asymmetry slopes are.

The record-level analysis gives the same conclusion. Across all self-verification decisions, the adjusted odds of verifier error are 56.2 times higher when the frozen candidate is incorrect (95\% CI $[51.1,61.9]$). This odds ratio is equivalent to $\operatorname{odds}(\mathrm{FPR})/
\operatorname{odds}(\mathrm{FNR})$, measuring the likelihood of the directional preference.

Together, these results establish the directional component of the Verification Mirage. Verification does not simply make frequent errors. Its errors are disproportionately concentrated on accepting incorrect candidates. The next analysis asks whether this failure is especially strong for errors
produced by the verifier itself.

\subsection{Preferential Acceptance of a Model's Own Errors}
\label{app:own_error_control}

Directional false acceptance does not show that generator and verifier errors are linked. A permissive verifier could accept many wrong answers regardless of which model produced them. We therefore ask a stricter question: among candidates that are already known to be wrong, does a verifier accept its own errors more often than errors produced by other models?

We build a complete generator-verifier matrix for each dataset. Every incorrect frozen candidate from each of the six generators is judged by all six verifiers under the same prompt. Generator identity is hidden from the verifier. Because every candidate in this analysis is incorrect, every acceptance is a false acceptance. Across five datasets, the matrix contains 122,364 judgments over 20,394 candidate records and 4,646 questions. All 180 dataset-generator-verifier settings are complete.

We fit the following binomial logistic mixed-effects model:
\[
\begin{aligned}
\texttt{accept} \sim\;&
\texttt{verifier}
+
\texttt{generator}
+
\texttt{dataset}
+
\texttt{verifier:self} \\
&+
(1\mid\texttt{question\_id})
+
(1\mid\texttt{input\_id})
+
(1\mid\texttt{question\_id:verifier}).
\end{aligned}
\]

Here, \texttt{self} indicates that generator and verifier are the same model. The verifier terms account for differences in general false-acceptance tendency. The generator terms account for differences in how readily each model's errors are accepted. The verifier-specific \texttt{self} term
therefore measures the additional odds of accepting a wrong answer when it
was produced by the verifier itself. We report Wald 95\% confidence intervals and apply Holm correction across the six verifier-specific tests.

As shown in Table~\ref{tab:model_specific_self_effects}, all six verifiers show a significant own-error excess. Adjusted odds ratios range from
2.32 to 8.53, with all Holm-adjusted $p<.001$. The corresponding model-dataset estimates are shown in Figure~\ref{fig:error_inheritance}.

\begin{table}[h]
\centering
\small
\caption{\textbf{Preferential acceptance of each verifier's own errors.} Adjusted odds ratios compare acceptance of a verifier's own incorrect candidates with incorrect candidates produced by other models after
accounting for verifier, generator, and dataset effects. Values above one
indicate greater acceptance of the verifier's own errors.}
\label{tab:model_specific_self_effects}
\begin{tabular}{@{}lrrr@{}}
\toprule
Verifier & OR & 95\% CI & Holm $p$ \\
\midrule
Qwen3-VL-8B         & 3.17 & [2.81, 3.58] & $<.001$ \\
Gemma-4-E4B         & 5.38 & [4.68, 6.19] & $<.001$ \\
Phi-4-Multimodal    & 2.37 & [2.11, 2.67] & $<.001$ \\
MedGemma-1.5-4B     & 8.53 & [7.37, 9.87] & $<.001$ \\
Lingshu-7B          & 2.32 & [2.06, 2.62] & $<.001$ \\
HuatuoGPT-Vision-7B & 4.36 & [3.63, 5.25] & $<.001$ \\
\bottomrule
\end{tabular}
\end{table}

As shown, this result goes beyond general verifier permissiveness. The comparison is
made after accounting for both the verifier's overall acceptance tendency and systematic differences among generators' errors. Even under these controls, every verifier is more likely to accept an error produced by itself. This is the primary evidence for the own-error dependence reported in the main paper.

The result does not identify the mechanism behind this dependence. Shared blind spots, answer style, or other model-specific properties may contribute. We therefore next test whether the effect can be explained more simply by a general preference for self-generated outputs.

\subsection{Is This Just General Self-Preference?}
\label{app:source_correctness}

The controlled analysis above shows that verifiers preferentially accept their own wrong answers. However, this does not yet establish that the effect is
specific to errors. A model may simply favor its own outputs regardless of whether they are correct. In this subsection, we therefore ask whether own-source preference is stronger for incorrect than for correct candidates.

We construct a correctness-balanced setting with equal numbers of correct and incorrect candidates from every generator-dataset stratum, yielding 1,200 frozen
candidate records. All six verifiers freshly judge every candidate under the same vanilla prompt, with generator identity hidden. Correctness labels are
frozen before verification. The resulting design contains 7,200 judgments and balances candidate correctness within every generator-dataset stratum.

For each verifier, we first measure its preference for its own outputs over outputs from other models, separately for correct and incorrect candidates.
We then ask whether this preference is stronger when the candidate is wrong. A positive wrong-minus-correct difference indicates stronger own-source
preference on errors. This analysis accounts for systematic differences across verifiers, generators, and datasets.

\begin{figure}[htbp]
\centering
\includegraphics[width=\linewidth]{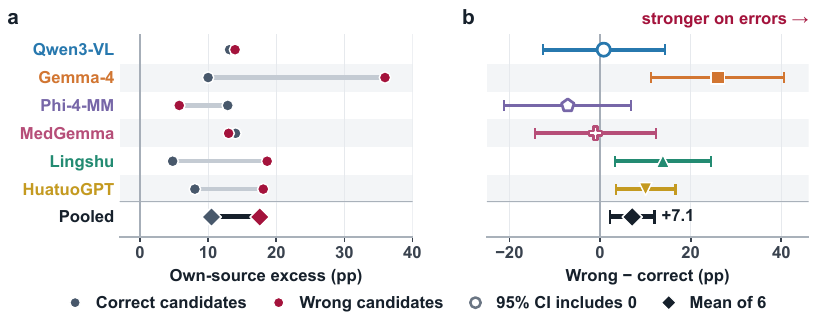}
\caption{\textbf{Correctness-conditioned own-source preference.} (a) Adjusted acceptance excess for each verifier's own outputs over outputs from other models, shown separately for correct and incorrect candidates. (b) The corresponding wrong-minus-correct difference. Positive values indicate stronger own-source preference when the candidate is wrong. Horizontal bars show 95\% confidence intervals, with shared images treated as clusters. Pooled estimates weight the six verifiers equally.}
\label{fig:source_interaction}
\end{figure}

As shown in Figure~\ref{fig:source_interaction}, Panel (a) shows that verifiers can prefer their own outputs for both correct and incorrect candidates. Panel (b) asks whether that preference is stronger
when the candidate is wrong. Averaged across the six verifiers, the wrong-minus-correct difference is $+7.09$ percentage points (95\% CI $[2.21,11.97]$, $p=.00439$). This indicates that own-source
preference is stronger for incorrect candidates on average.

\begin{table}[htbp]
\centering
\small
\setlength{\tabcolsep}{4pt}
\caption{\textbf{Correctness-conditioned source preference by verifier.} Wrong and correct excess report adjusted own-versus-other acceptance differences. Their difference is the wrong-minus-correct contrast shown in Figure~\ref{fig:source_interaction}b. Adjusted odds ratios report the corresponding multiplicative interaction with Holm-corrected $p$ values.}
\label{tab:source_correctness_interaction}
\begin{tabular}{@{}lrrrrr@{}}
\toprule
Model & Wrong excess & Correct excess & Difference [95\% CI] & Adj. OR & $p_H$ \\
\midrule
Qwen3-VL & 13.95 & 13.14 & 0.80 [-12.66, 14.27] & 0.94 & 1 \\
Gemma-4 & 35.96 & 9.99 & 25.97 [11.30, 40.65] & 3.52 & 0.0899 \\
Phi-4-MM & 5.75 & 12.87 & -7.12 [-21.13, 6.89] & 0.72 & 1 \\
MedGemma & 13.00 & 14.02 & -1.01 [-14.41, 12.39] & 0.82 & 1 \\
Lingshu & 18.66 & 4.79 & 13.88 [3.27, 24.49] & 1.66 & 0.86 \\
Huatuo & 18.09 & 8.05 & 10.04 [3.46, 16.62] & 0.99 & 1 \\
\bottomrule
\end{tabular}
\end{table}

These results distinguish own-error dependence from a general preference for self-generated outputs. Verifiers can favor their own answers even when those
answers are correct, but this preference is stronger on incorrect answers on average. Generic self-preference therefore cannot fully account for the own-error effect in Section~\ref{app:own_error_control}.

\section{Mechanism Diagnostics}
\label{app:mechanism_diagnostics}
Section~\ref{sec:simpler_explanations} points to a gap between having access to the image and obtaining evidence that can actually test a candidate. \emph{Here we examine that gap under controlled interventions and provide new outcomes to supplement the study in main paper}. We first ask whether stronger access to the visual input is sufficient for discrimination. We then remove direct candidate conditioning and ask whether an independent solution escapes the original error. Finally, we test whether the useful ingredient is simply more correct visual information, or information that directly distinguishes the candidate from an alternative.

The image-replacement and candidate-hidden re-solving analyses use the same diagnostic setup. For each of the 30 model-dataset settings, we sample balanced correct and incorrect answers. The image, question, candidate, prompt, and decoding settings are held fixed except for the intervention under study. Reference answers are used
only after inference for correctness stratification. The trusted-region and controlled spatial analyses use annotation-constrained SLAKE subsets described below. We note that \verimap{} audits the whole datasets in the main text, but we only sample data for further diagnostics here for efficiency.

\subsection{Visual Access Does Not Guarantee Discrimination}
\label{app:visual_access}

\paragraph{Image dependence is present, but only weakly selective.}
We first test whether the Verification Mirage can be explained by insufficient dependence on the current image. We replace the original image with its nearest nonidentical neighbor under BiomedCLIP cosine similarity, restricting the search to the same dataset and task stratum~\citep{zhang2023biomedclip}. When no eligible image exists within the stratum, we fall back to the same dataset, which occurs in 17 of 3,000 cases.

As shown in Table~\ref{tab:image_intervention_inference},
we evaluate candidates that were accepted under the original image and ask whether that acceptance survives the replacement. Among 1,055 originally false acceptances, 784 remain accepted, corresponding to 74.3\%. Among 1,218
true acceptances, 970 remain accepted, corresponding to 79.6\%. The resulting correct-minus-incorrect retention gap is 5.33 percentage points (95\% image-cluster bootstrap CI [0.75, 9.93], $p=.023$). Excluding the
fallback cases gives a similar gap of 5.23 points.

Thus, verifier decisions do respond to image content, but the response is only weakly selective in this intervention. Nearly three quarters of the original false acceptances survive the image change. \emph{We therefore do not interpret this result as evidence that the verifier ignores the image.} The nearest replacement may also preserve the relevant finding. Instead, the experiment shows that
dependence on the current image does not by itself provide a strong signal for separating correct from incorrect acceptances.

\begin{table}[htbp]
\centering
\small
\setlength{\tabcolsep}{4pt}
\setlength{\belowcaptionskip}{5pt}
\caption{\textbf{Acceptance after image replacement.}
Among 2,273 originally accepted candidates, the original image is replaced by its nearest nonidentical neighbor. Intervals use 5,000 image-cluster bootstrap resamples.}
\label{tab:image_intervention_inference}
\begin{tabular}{@{}lrr@{}}
\toprule
Initial candidate & Retained / accepted & Retention [95\% CI] \\
\midrule
Incorrect & 784/1055 & 74.31 [70.83, 77.64] \\
Correct & 970/1218 & 79.64 [76.22, 82.67] \\
Selectivity gap & NA & 5.33 [0.75, 9.93] \\
Gap without fallback & NA & 5.23 [0.65, 9.85] \\
\bottomrule
\end{tabular}
\end{table}

\paragraph{Even trusted access to the relevant region does not reliably improve discrimination.}
Image replacement is an indirect test of visual dependence. A stronger possibility is that the verifier has access to the correct image but fails to locate the relevant content. We test this explanation using SLAKE native bounding boxes, which define a reference-assisted offline region of interest
for 194 eligible model-question pairs spanning 42 image clusters~\citep{liu2021slake}. We either mark the trusted region on the original image or crop the image to that region.

Marking the region reduces acceptance by 7.63 points for incorrect candidates and 6.35 points for correct candidates. The resulting change in discrimination, defined as correct-candidate acceptance minus incorrect-candidate acceptance, is +1.28 points (95\% image-cluster CI [-7.95, 10.99]). Cropping reduces acceptance by 13.74 points for incorrect candidates and 23.81 points for correct candidates, giving a discrimination change of -10.07 points (CI [-24.86, 4.34]). Neither intervention establishes improved discrimination after Holm correction, with $p=.789$ for marking and $p=.355$ for cropping.

Together, these results separate visual access from discriminative evidence. We do not find statistically meaningful evidence that providing the correct image or the relevant region improves the separation between correct and incorrect candidates. Verification therefore requires more than locating the relevant pixels. It requires an observation that can distinguish the proposed answer from its alternatives.

\begin{table}[htbp]
\centering
\small
\setlength{\tabcolsep}{4pt}
\setlength{\belowcaptionskip}{5pt}
\caption{\textbf{Trusted-region access does not reliably improve discrimination.}
Results are paired changes in acceptance on 194 eligible SLAKE model-question
pairs. Discrimination change is the change for correct candidates minus the
change for incorrect candidates. Positive values are favorable. Values are
percentage points with image-cluster 95\% intervals.}
\label{tab:roi_inference}
\begin{tabular}{@{}lrrr@{}}
\toprule
Intervention & $\Delta$ false accept & $\Delta$ true accept & $\Delta$ discrimination \\
\midrule
Marked & -7.63 [-16.05, -0.84] & -6.35 [-13.11, -1.45] & 1.28 [-7.95, 10.99] \\
Crop & -13.74 [-21.21, -6.34] & -23.81 [-36.24, -11.76] & -10.07 [-24.86, 4.34] \\
\bottomrule
\end{tabular}
\end{table}

\subsection{From Independent Re-Solving to a Falsifiable Check}
\label{app:independent_resolving}

\paragraph{Candidate hiding rarely repairs the original error.}
The previous analyses ask whether stronger visual access is sufficient. We next test the second natural explanation from
Section~\ref{sec:simpler_explanations}: verification may fail because the proposed answer directly anchors the verifier. If candidate conditioning were the main obstacle, hiding the candidate and solving the image-question pair again should often recover the correct answer.

We therefore independently rescore both the frozen candidate and its candidate-hidden re-solution after inference. Across 1,503 originally incorrect diagnostic candidates, only 125 re-solutions are correct, corresponding to 8.32\% (95\% image-cluster bootstrap CI [6.85, 9.95]). Among the remaining cases, 1,138, or 75.7\%, are classified by the original semantic-relation annotation as repeating or remaining compatible with the same error. Another 240, or 16.0\%, produce a different or unresolved error.

To verify that this low correction rate is not driven by the original correctness labels, we rescore both the frozen candidates and their candidate-hidden re-solutions with the same fresh semantic judge. Only five of the 1,503 original candidates change from incorrect to correct under this rescoring. Among the remaining 1,498 candidates judged incorrect by both procedures, re-solving corrects only 120 cases, or 8.01\% (95\% CI [6.62, 9.58]). Correction also remains limited for both binary-answer questions (13.9\%) and open-answer questions (7.3\%), indicating that the low recovery rate is not driven by a single question format.

The 79.0\% result in Figure~\ref{fig:existing_explanations}b measures whether candidate-hidden re-solving reproduces an answer that is semantically compatible with the original error. The analysis above asks a complementary
and stricter question: \emph{does re-solving actually recover the correct answer?} Both results point to the same conclusion. Hiding the candidate often leaves
the underlying error unresolved and rarely provides a reliable corrective signal.

\begin{table}[htbp]
\centering
\small
\setlength{\tabcolsep}{4pt}
\setlength{\belowcaptionskip}{5pt}
\caption{\textbf{Outcomes of candidate-hidden re-solving.}
Both the frozen candidate and the re-solved answer are independently rescored
after inference. Correction uses the fresh correctness labels. Same-error
identity uses the original semantic-relation annotation. Values are
percentages with 95\% image-cluster bootstrap intervals.}
\label{tab:resolving_outcomes}
\begin{tabular}{@{}lrrrr@{}}
\toprule
Cohort & $n$ & Corrected & Same error & Other / unresolved \\
\midrule
Frozen wrong & 1503 & 8.32 [6.85, 9.95] & 75.72 [73.44, 78.08] & 15.97 [13.96, 17.99] \\
Both judges wrong & 1498 & 8.01 [6.62, 9.58] & 75.97 [73.71, 78.30] & 16.02 [14.01, 18.03] \\
\bottomrule
\end{tabular}
\end{table}

\paragraph{Disagreement becomes useful when it defines what to test rather than acting as a second vote.}
Candidate-hidden re-solving removes direct access to the proposed answer, but it still asks the model to solve essentially the same visual problem again. We therefore change the role of the second answer. When a candidate-hidden answer disagrees with an accepted candidate, the pair is used only to define the smallest visual distinction between the two hypotheses. A fresh observer then measures that distinction without seeing either answer or knowing which
one is the original candidate. The observation is frozen before a role-blind reconstruction step maps it back to an anonymous hypothesis.

The competing answer therefore specifies what the image must distinguish, but
it is not trusted as evidence and does not act as a vote. The diagnostic signature is selectivity. A useful check should revoke incorrect acceptances more often than it revokes correct ones. In the frozen probe summarized in Figure~\ref{fig:existing_explanations}c, this occurs in 25 of 30 model-dataset settings. Across all settings, the probe catches 23.9\% of false acceptances while harming 12.1\% of true acceptances.

This contrast identifies a more specific requirement than candidate independence alone. A second solution can reproduce the original failure. Disagreement becomes useful when it is converted into a visual test whose observed outcome determines which hypothesis remains supported. This probe
motivates the information flow used by \ours{}.

\subsection{Relevant Evidence Matters More Than Extra Information}
\label{app:fact_specificity}

The preceding falsification probe suggests that targeted evidence can make verification more selective. A remaining question is whether this benefit comes simply from giving the verifier more correct visual information. We isolate this possibility using 40 distinct SLAKE images with one unambiguous native bounding box per image.

\paragraph{A controlled probe separates discriminative evidence from additional information.} For each image, we construct questions asking whether the box center lies to the left or right of the image midpoint, or above or below it. Two opposite candidate answers balance correctness. Five verifiers receive one of three inputs: no additional fact, the true coordinate on the orthogonal axis, or the true coordinate on the axis needed to answer the question. Images and candidates are identical across conditions, giving 400 candidate-model cases per condition and 1,200 calls in total at temperature zero. All selected images are square, so resizing preserves coordinate fractions.

Both supplied coordinates are correct annotation facts. The difference is whether the fact can resolve the question being asked. For a left-versus-right question, for example, the horizontal coordinate is discriminative while the vertical coordinate is not. This design therefore separates the value of answer-relevant evidence from the value of providing additional correct information.

\begin{table}[htbp]
\centering
\small
\setlength{\tabcolsep}{4pt}
\setlength{\belowcaptionskip}{5pt}
\caption{\textbf{Verification with visual facts.} SLAKE images are used to construct horizontal and vertical box-center questions with two opposing candidate answers. Each condition includes
400 matched candidate-model cases across five verifiers. Yield is the fraction of all 400 attempts that return a parseable and correct verdict. Binary Acc. is computed over parseable verdicts only. Supplied facts are derived from native bounding-box annotations.}
\label{tab:evidence_specificity}
\begin{tabular}{@{}lrrr@{}}
\toprule
Provided fact & Explicit verdicts & Yield [95\% CI] & Binary Acc. [95\% CI] \\
\midrule
None & 355/400 & 48.25 [43.75, 52.26] & 54.80 [49.87, 59.34] \\
Orthogonal coordinate & 332/400 & 44.25 [40.00, 48.75] & 52.88 [48.70, 57.00] \\
Requested coordinate & 379/400 & 57.25 [54.00, 60.25] & 60.42 [57.44, 63.16] \\
\bottomrule
\end{tabular}
\end{table}

\paragraph{Only the discriminative fact produces a clear improvement.}
Because output availability differs across conditions, we also report correct-verdict yield per attempt. This exploratory outcome was chosen after inspecting output availability.

\begin{figure}[htbp]
\centering
\includegraphics[width=\linewidth]{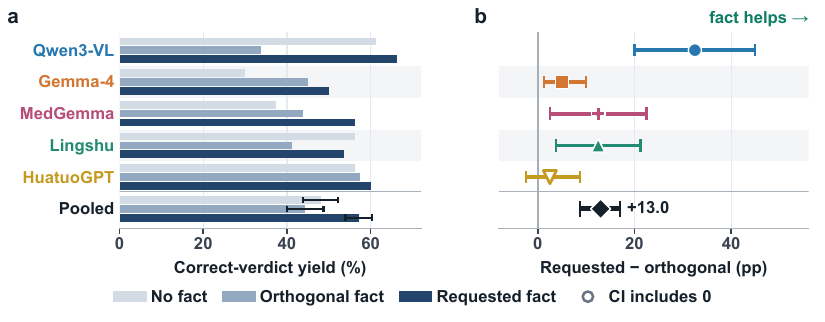}
\caption{\textbf{Discriminative visual facts improve verification.}
(a) Correct-verdict yield for each verifier and the pooled result with no additional fact, a correct but non-informative orthogonal-axis fact, or the requested-axis fact needed to answer the question. (b) Paired improvement from the requested-axis fact over the orthogonal-axis fact for each verifier and in the pooled result. Each verifier is evaluated on 80 paired candidate cases from 40 images, and the pooled estimate weights verifiers equally. Error bars show 95\% image-cluster bootstrap confidence intervals. Hollow markers in panel (b) indicate intervals that include zero. All supplied facts come from native bounding-box annotations.}
\label{fig:fact_specificity}
\end{figure}

As shown in Table~\ref{tab:evidence_specificity} and Figure~\ref{fig:fact_specificity}a, correct-verdict yield is 48.25\% with no additional fact, 44.25\% with the orthogonal coordinate, and 57.25\% with the
requested coordinate. Providing the requested-axis fact improves yield over
the orthogonal-axis fact by 13.00 percentage points
(95\% paired image-cluster bootstrap CI [8.75, 17.00]). It also improves
yield over providing no fact by 9.00 points
(CI [4.75, 13.25]). Both contrasts remain significant after Holm correction
with $p<10^{-4}$.

We also test whether differing output availability could explain the result.
Unavailable outputs occur in 4 cases with no fact, 18 cases with the
orthogonal fact, and 21 cases with the requested fact. Assigning unavailable
outcomes adversarially still leaves positive requested-axis advantages of
[8.50, 18.25] points over the orthogonal fact and [8.00, 14.25] points over
no fact. 

Together, these results provide a limited sufficiency test. Additional correct
visual information does not automatically improve verification. The benefit
appears when the supplied fact directly resolves the distinction on which the
candidate's correctness depends. Because these facts come from privileged
annotations, this experiment does not establish that the deployed
\textsc{Measure} stage can always recover the required evidence.

Taken together, the mechanism diagnostics narrow the missing ingredient beyond
visual access or candidate independence alone. Verification becomes more
selective when the available evidence directly resolves the visual distinction
that determines whether the candidate is correct. Appendix~E operationalizes
this criterion in \ours{} by separating contrast construction, visual
measurement, and evidence-to-verdict binding.

\section{\ours{}: Method Details}
\label{app:miragecheck_spec}

Appendix~\ref{app:mechanism_diagnostics} identifies two requirements for a discriminative visual check. First, a competing answer should define what the image must distinguish rather than act as a second vote. Second, the relevant visual evidence should be formed before the verifier knows which hypothesis is the original candidate. \ours{} implements these requirements through \textsc{Contrast}, \textsc{Measure}, and \textsc{Bind}.

\subsection{Protocol}

\paragraph{From disagreement to an evidence-bound decision.}
\ours{} is applied only to candidates initially accepted by the verifier. Let $(x,q,a)$ denote the image, question, and frozen candidate. The model first answers the image-question pair in a fresh context without seeing $a$, producing a rival answer $r$.

In \textsc{Contrast}, the disagreement between $a$ and $r$ is converted into a neutral visual question that distinguishes the two hypotheses. The image is hidden at this stage, so \textsc{Contrast} specifies what should be checked rather than answering the check itself. In \textsc{Measure}, a fresh observer receives only the image and this visual question. Neither answer is visible. The observer records the relevant visual evidence, which is then frozen.

In \textsc{Bind}, the frozen evidence is shown together with the two answers in
randomized anonymous slots. The model selects the answer best supported by the evidence without knowing which slot contains the original candidate. The initial acceptance is retained only if the selected answer maps back to $a$. Otherwise, it is revoked.

\paragraph{The rival defines the test, not the verdict.}
All calls within a model-dataset setting use the same frozen VLM checkpoint in fresh contexts. Reference answers, post-hoc correctness labels, and generator
identity are never available during inference. The rival answer is used only
to expose a disagreement that can be tested visually. It is not treated as
evidence and does not vote against the candidate. Visual evidence is formed
before either answer is revealed to the observer, and candidate identity
remains hidden while that evidence is matched back to the two hypotheses.

\subsection{What Each Stage Can See}

The three stages intentionally receive different information
(Table~\ref{tab:miragecheck_information}). \textsc{Contrast} sees the competing
answers but not the image, allowing it to define the distinction that needs to
be tested without producing the visual evidence itself. \textsc{Measure} sees
the image and the registered visual question but not either answer, preventing
the candidate or rival from directly shaping the observation.
\textsc{Bind} receives the frozen evidence and the two answers in anonymous
slots, but does not know which answer is the original candidate. Candidate
identity is restored only after the evidence-to-answer match has been made.

This separation gives each stage a single role. \textsc{Contrast} determines
what to check, \textsc{Measure} determines what the image shows, and
\textsc{Bind} determines which hypothesis is supported by that observation.
The protocol therefore prevents the rival from becoming a second vote and
prevents candidate identity from directly influencing visual evidence
formation.

\begin{table}[t]
\centering
\small
\setlength{\tabcolsep}{4.5pt}
\renewcommand{\arraystretch}{1.08}
\caption{\textbf{Information barriers in \ours{}.}
Each stage receives only the information required for its role. The competing
answers are hidden while visual evidence is formed, and candidate identity is
hidden while that evidence is matched back to the hypotheses.}
\label{tab:miragecheck_information}
\begin{tabular}{@{}p{0.14\linewidth}p{0.28\linewidth}p{0.25\linewidth}p{0.25\linewidth}@{}}
\toprule
Stage & Purpose & Available information & Hidden information \\
\midrule
\textsc{Contrast}
& Define what the image must distinguish
& Candidate and rival
& Image \\

\textsc{Measure}
& Observe the relevant visual evidence
& Image and visual question
& Candidate and rival \\

\textsc{Bind}
& Match the frozen evidence to a hypothesis
& Frozen evidence and anonymous answers
& Candidate identity \\
\bottomrule
\end{tabular}
\end{table}




\section{Additional \ours{} Results and Controls}
\label{app:miragecheck_results}

\subsection{Evaluation Coverage}
\label{app:miragecheck_cohorts}

\paragraph{Evaluation coverage.}
All five datasets are used exclusively for evaluation. All methods operate on frozen source candidates, and correctness labels are assigned post hoc. The full evaluation comprises 30 model-dataset settings across six VLMs and five medical VQA benchmarks. Some auxiliary analyses and ablations are conducted on subsets of these datasets for computational efficiency or because particular annotations are required, as specified in the corresponding sections.



\paragraph{Pairing and inference cost.}
Method comparisons use matched source records whenever the corresponding outputs are available. Failed calls are tracked separately. On the fully recorded six-model VQA-RAD and SLAKE dataset, the implemented \ours{} pipeline requires 9.17 model requests per sample on average.

\subsection{Results Across Models and Datasets}
\label{app:miragecheck_consistency}

\paragraph{Performance across models.}
Table~\ref{tab:miragecheck_by_model} reports five-dataset macros by model. Relative to SGV, \ours{} improves macro BAcc and reduces macro FPR for all six models. Effect sizes vary substantially across checkpoints, so the aggregate gain is not driven by one model.

\begin{table}[!htbp]
\centering
\caption{\textbf{Per-model comparison across verification methods.} Results
are macro-averaged over five datasets.}
\label{tab:miragecheck_by_model}
\setlength{\tabcolsep}{1.5pt}
\renewcommand{\arraystretch}{0.97}
\scriptsize
\begin{tabular}{@{}lcccccccccccc@{}}
\toprule
 & \multicolumn{2}{c}{Qwen3-VL} & \multicolumn{2}{c}{Gemma-4} & \multicolumn{2}{c}{Phi-4-MM} & \multicolumn{2}{c}{Lingshu} & \multicolumn{2}{c}{Huatuo} & \multicolumn{2}{c}{MedGemma} \\
\cmidrule(lr){2-3} \cmidrule(lr){4-5} \cmidrule(lr){6-7} \cmidrule(lr){8-9} \cmidrule(lr){10-11} \cmidrule(lr){12-13}
Method & BAcc\,$\uparrow$ & FPR\,$\downarrow$ & BAcc\,$\uparrow$ & FPR\,$\downarrow$ & BAcc\,$\uparrow$ & FPR\,$\downarrow$ & BAcc\,$\uparrow$ & FPR\,$\downarrow$ & BAcc\,$\uparrow$ & FPR\,$\downarrow$ & BAcc\,$\uparrow$ & FPR\,$\downarrow$ \\
\midrule
Vanilla & 56.90 & 56.47 & 53.21 & 86.92 & 58.81 & 46.35 & 56.71 & 66.81 & 52.67 & 91.89 & 52.88 & 77.36 \\
Exact Second Look & 51.57 & 94.14 & 50.17 & 93.80 & 58.54 & 64.82 & 51.65 & 95.73 & 51.13 & 95.88 & 50.33 & 95.00 \\
SGV & 61.24 & 39.81 & 54.52 & 82.44 & 56.50 & 58.40 & 61.37 & 42.26 & 52.20 & 80.83 & 57.83 & 52.63 \\
\cdashline{1-13}[0.35pt/2pt]\addlinespace[1.5pt]
CausalLens & 58.41 & 45.51 & 53.26 & 86.84 & 55.63 & \textbf{29.75} & 54.94 & 57.09 & 52.30 & 88.44 & 52.87 & 78.76 \\
ClearSight & 54.42 & 68.22 & 53.45 & 86.56 & 56.54 & 36.12 & 55.03 & 64.58 & 52.83 & 92.07 & 53.49 & 71.89 \\
\cdashline{1-13}[0.35pt/2pt]\addlinespace[1.5pt]
\rowcolor{resultblue!20}\ours{} & \textbf{62.51} & \textbf{29.11} & \textbf{61.50} & \textbf{55.28} & \textbf{59.66} & 37.64 & \textbf{62.21} & \textbf{27.69} & \textbf{55.21} & \textbf{67.04} & \textbf{57.98} & \textbf{51.83} \\
\midrule
 & \multicolumn{2}{c}{Qwen3-VL} & \multicolumn{2}{c}{Gemma-4} & \multicolumn{2}{c}{Phi-4-MM} & \multicolumn{2}{c}{Lingshu} & \multicolumn{2}{c}{Huatuo} & \multicolumn{2}{c}{MedGemma} \\
\cmidrule(lr){2-3} \cmidrule(lr){4-5} \cmidrule(lr){6-7} \cmidrule(lr){8-9} \cmidrule(lr){10-11} \cmidrule(lr){12-13}
Method & $|$Bias$|$\,$\downarrow$ & D-Skew\,$\downarrow$ & $|$Bias$|$\,$\downarrow$ & D-Skew\,$\downarrow$ & $|$Bias$|$\,$\downarrow$ & D-Skew\,$\downarrow$ & $|$Bias$|$\,$\downarrow$ & D-Skew\,$\downarrow$ & $|$Bias$|$\,$\downarrow$ & D-Skew\,$\downarrow$ & $|$Bias$|$\,$\downarrow$ & D-Skew\,$\downarrow$ \\
\midrule
Vanilla & 23.98 & 17.67 & 51.65 & 49.83 & 18.01 & 12.80 & 26.59 & 20.21 & 49.51 & 48.07 & 38.38 & 32.73 \\
Exact Second Look & 51.87 & 51.03 & 56.66 & 55.82 & 35.87 & 30.56 & 50.03 & 49.71 & 51.85 & 51.07 & 53.15 & 51.91 \\
SGV & 9.73 & 4.01 & 48.69 & 46.31 & 27.67 & 20.16 & 10.02 & 4.13 & 38.42 & 33.89 & 20.77 & 16.52 \\
\cdashline{1-13}[0.35pt/2pt]\addlinespace[1.5pt]
CausalLens & 18.34 & 11.27 & 51.52 & 49.80 & 13.70 & 6.94 & 15.17 & 9.77 & 45.83 & 43.13 & 39.67 & 34.22 \\
ClearSight & 29.57 & 24.22 & 51.46 & 49.65 & 13.05 & 6.00 & 22.89 & 16.60 & 49.88 & 48.65 & 33.49 & 27.34 \\
\cdashline{1-13}[0.35pt/2pt]\addlinespace[1.5pt]
\rowcolor{resultblue!20}\ours{} & \textbf{5.88} & \textbf{1.72} & \textbf{27.68} & \textbf{21.60} & \textbf{9.54} & \textbf{3.96} & \textbf{9.04} & \textbf{4.04} & \textbf{27.62} & \textbf{20.35} & \textbf{20.10} & \textbf{15.93} \\
\bottomrule
\end{tabular}
\end{table}

\paragraph{Setting-level consistency and exceptions.}
Table~\ref{tab:miragecheck_cell_consistency} summarizes paired comparisons with SGV in the preferred direction across the 30 model-dataset settings. BAcc improves in 23 settings, ties for MedGemma on VQA-RAD, and declines for five models on MedXpert-QA and Lingshu on SLAKE. FPR improves in 29 settings and ties in one. Thus, the broad reduction in false acceptance coexists with specific BAcc regressions across the evaluation matrix.

\begin{table}[htbp]
\centering
\small
\caption{\textbf{Setting-level consistency of \ours{} versus SGV.} Counts compare the 30 model-dataset settings in each metric's preferred direction, using the reported two-decimal setting values.}
\label{tab:miragecheck_cell_consistency}
\setlength{\tabcolsep}{10pt}
\begin{tabular}{@{}lrrr@{}}
\toprule
Metric & Better & Tie & Worse \\
\midrule
BAcc & 23 & 1 & 6 \\
FPR & 29 & 1 & 0 \\
$|\mathrm{Bias}|$ & 25 & 1 & 4 \\
D-Skew & 25 & 1 & 4 \\
\bottomrule
\end{tabular}
\end{table}

\subsection{Comparison with Simple Rejection Controls}
\label{app:rejection_selectivity}

We test whether the improvement from \ours{} can be explained by simply
rejecting more initial acceptances or by using disagreement from one
candidate-hidden re-solve.

\paragraph{Matched random rejection.}
As a control for rejection alone, we randomly revoke initial acceptances
using the same rejection budget as \ours{}. This reduces FPR relative to
Vanilla, but increases FNR enough that BAcc decreases from 56.36\% to
55.32\% (Table~\ref{tab:rejection_controls}). Thus, the gain from
\ours{} cannot be explained by a more conservative acceptance rate alone.

\paragraph{Candidate-hidden re-solve gate.}
We next test a stronger baseline that uses only candidate-independent
re-solving. For each initially accepted candidate, the same VLM
re-answers the original image--question pair once at temperature zero
using the original VQA prompt, without access to the frozen candidate.
The initial acceptance is retained when the re-solved answer is
semantically consistent with the candidate and revoked otherwise.
Initial rejections remain unchanged. This baseline requires no
\textsc{Contrast}, \textsc{Measure}, or \textsc{Bind} stage.

The re-solve gate improves BAcc from 56.36\% to 58.24\%, showing that
candidate-hidden disagreement itself provides a useful error signal.
However, Full \ours{} reaches 64.15\% BAcc, a further 5.91-point gain,
while reducing FPR from 64.86\% to 40.29\%. The accompanying FNR
increase from 18.66\% to 31.41\% is smaller than the FPR reduction, so
the resulting tradeoff remains favorable in balanced accuracy. These
results show that one-shot candidate-hidden disagreement explains only
part of the improvement, while the complete visual recheck provides a
substantial additional gain.

\begin{table}[h!]
\centering
\small
\setlength{\belowcaptionskip}{5pt}
\caption{\textbf{Comparison with simple rejection controls on PMC-VQA.}
Random rejection matches the rejection budget of \ours{}. The
candidate-hidden re-solve gate re-answers the original image--question
pair once without the candidate and revokes an initial acceptance when
the two answers disagree.}
\label{tab:rejection_controls}
\begin{tabular}{@{}lccc@{}}
\toprule
Method & BAcc $\uparrow$ & FPR $\downarrow$ & FNR $\downarrow$ \\
\midrule
Vanilla
    & 56.36 & 72.47 & \textbf{14.81} \\
Matched random rejection
    & 55.32 & 66.27 & 23.09 \\
Candidate-Hidden Re-solve Gate
    & 58.24 & 64.86 & 18.66 \\
Full \ours{}
    & \textbf{64.15} & \textbf{40.29} & 31.41 \\
\bottomrule
\end{tabular}
\end{table}

\subsection{Visual Evidence vs. a Second Vote}
\label{app:evidence_decision_controls}

\paragraph{Candidate-role sensitivity.}
For Phi-4, 255 initially accepted pairs with distinct hypotheses were rerun while only the proposed-answer role was swapped; the image, question, and candidate-hidden prior stayed fixed. The initial screen accepted both answers in 187/255 cases (73.3\%; Wilson 95\% CI [67.6, 78.4]). Among explicit yes/no opposites, dual acceptance reached 63/68 (92.6\%; CI [83.9, 96.8]). Candidate-visible judging can therefore endorse incompatible hypotheses, motivating role-blind binding; this result does not establish that \textsc{Bind} selects the correct answer.

\paragraph{Evidence-guided selection versus a second vote.}
On the frozen six-model VQA-RAD/SLAKE evaluation subset, treating the candidate-hidden challenger as a vote leaves Accuracy essentially unchanged (63.02\% to 63.01\%). Using the recorded visual observation and independent trace for evidence-guided selection raises Accuracy to 64.28\% and reduces FPR from 63.24\% to 48.28\%. The rival is useful for defining what to check when the resulting evidence governs selection; another vote does not yield the same benefit. Because the selector also sees the independent trace, this comparison does not independently isolate the targeted observation.

\subsection{Ablating Information Hiding}
\label{app:information_ablation}

The stage-deletion analysis in Table~\ref{tab:miragecheck_ablation} shows
that contrast construction alone is insufficient: the full evidence-bound recheck is required to recover the observed gains. To more directly test the
information-separation design, we run two functional ablations on the same frozen six-model PathVQA cohort. Frozen source candidates, initial decisions, rival answers, \textsc{Contrast} checks, randomized hypothesis slots, decoding settings, and post-hoc labels are held fixed. The altered stages are reached in 361 cases. All other decisions are
replayed unchanged. In \emph{Answer-visible Measure}, the observer additionally sees the two role-unlabeled competing hypotheses while forming the visual observation. In \emph{Role-visible Bind}, the original candidate-hidden observation is reused, but the original-candidate role is revealed during
binding. A contemporaneous Full replay controls for inference-time variation.

\begin{table}[htbp]
\centering
\small
\caption{\textbf{Functional information-separation ablations on PathVQA.}
Values are six-model macros (\%).}
\label{tab:information_ablation}
\setlength{\tabcolsep}{5pt}
\begin{tabular}{@{}lrrrr@{}}
\toprule
Variant &
Acc. $\uparrow$ &
FPR $\downarrow$ &
$|$Bias$|$ $\downarrow$ &
D-Skew $\downarrow$ \\
\midrule
Historical Full
& 54.61 & 52.98 & 20.08 & 14.72 \\
Contemporaneous Full replay
& 54.91 & 52.79 & 20.07 & 14.75 \\
Answer-visible \textsc{Measure}
& 53.59 & 55.83 & 22.29 & 17.02 \\
Role-visible \textsc{Bind}
& 54.15 & 53.93 & 20.47 & 15.22 \\
\bottomrule
\end{tabular}
\end{table}

Revealing the competing hypotheses during \textsc{Measure} increases FPR by
3.04 percentage points relative to the contemporaneous Full replay
(95\% image-cluster bootstrap CI $[0.61,5.55]$) and decreases Accuracy by
1.32 points (CI $[-2.87,0.19]$), supporting the value of forming visual
evidence before exposing the competing answers. Revealing candidate identity
during \textsc{Bind} produces a smaller change (FPR $+1.14$ points,
CI $[-0.40,2.63]$; Accuracy $-0.76$ points, CI $[-1.67,0.17]$).
The historical and contemporaneous Full results differ by only $0.30$ points
in Accuracy and $0.19$ points in FPR, indicating limited replay variation.

\paragraph{Controls with matched model request counts.}
To test whether additional model calls explain the gains, we compare \ours{} with majority vote over $k$ vanilla verifications and $k$-sample re-solving
with an agreement threshold. We use $k=9$ at temperature zero on 200 frozen candidates per model and dataset across six models on MedXpert-QA and PMC-VQA. Nine requests approximate the 9.17-request mean reported above. Among the 1,103 of 2,400 cases with verdicts from all three methods, the 12-cell macro
Accuracy is 54.28\%, 45.35\%, and 31.99\% in that order. FPR is 55.79\%, 75.95\%, and 99.56\%, while FNR is 24.76\%, 10.82\%, and 1.25\%.
Corresponding verdict coverage is 2,246/2,400, 2,034/2,400, and 1,364/2,400. Additional requests alone do not reproduce \ours{}'s lower false-acceptance rate on this setting.

\section{Iterative Reuse and Self-Correction}
\label{app:iterative_reuse}

\subsection{Self-Improving Feedback Loop Setup}

\paragraph{The current-suite stress test tracks whether false acceptance becomes lock-in.}
The loop begins from the same frozen generator answer. At each turn, a fresh
verifier judges the current answer; its verdict and explanation are then passed
to the next generator context. The matched cohort contains all 696 PathVQA and
VQA-RAD trajectories for each current checkpoint, with five turns retained for
every model.

\subsection{False Acceptance Accumulates Across Turns}

\begin{figure}[htbp]
    \centering
    \includegraphics[width=\linewidth]{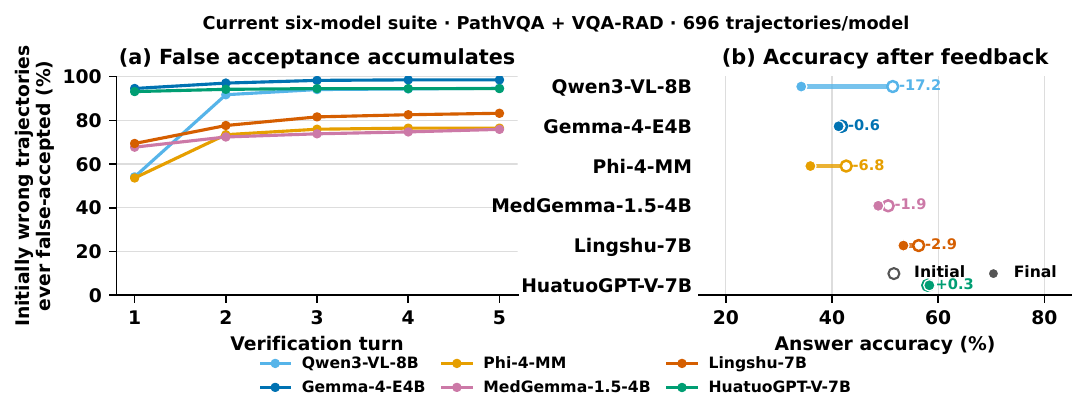}
    \caption{\textbf{Multi-turn feedback in the current six-model suite.}
    \textbf{(a)} Among initially incorrect trajectories, cumulative false
    acceptance is the fraction that has received at least one
    \textsc{Correct} verdict on an incorrect answer by the indicated turn.
    \textbf{(b)} Open and filled markers show initial and final answer
    accuracy; labels report percentage-point change. Early-stopped trajectories
    retain their cumulative event status.}
    \label{fig:feedback_loop_current}
\end{figure}

\paragraph{False acceptance accumulates under repeated reuse.}
By turn five, 75.9-98.5\% of initially incorrect trajectories have received at
least one false acceptance. Final accuracy decreases for five of six models;
the largest decline is 17.2 points for Qwen3-VL-8B, while
HuatuoGPT-Vision-7B changes by 0.3 points. This is a bounded two-dataset stress
test rather than evidence that every iterative feedback system degrades.

\subsection{MirageCheck in the Feedback Loop}

\paragraph{Feedback-loop cohort and setting-level exceptions.}
The implementation reported in
Section~\ref{sec:agent_repair} compares Vanilla and \ours{} on 6,000
trajectories (200 per model-dataset setting), initialized from the same frozen
answers. Table~\ref{tab:agent_repair_cells} reports every setting. Final
accuracy improves in 21 settings, ties in two, and falls in seven. Cumulative
false acceptance among initially incorrect trajectories falls in 27 settings
and rises in three, all for MedGemma. The averaged final accuracies are
38.58\% and 41.27\% and averaged cumulative false acceptance is 87.67\% and
42.72\%, respectively. The latter metric includes all initially incorrect trajectories in the denominator.

\begin{table}[htbp]
\centering
\footnotesize
\caption{\textbf{Complete matched feedback results.} Each entry is based on 200
frozen trajectories per method and setting. Van. = Vanilla, MC = \ours{}. Top: final answer accuracy (\%). Bottom: CFA, the
cumulative fraction of initially incorrect trajectories receiving at least
one false acceptance over five verifier audits (\%). Lower CFA is favorable. Average is computed over all 6,000 trajectories per method (30 settings $\times$ 200), so CFA is weighted by the number of initially incorrect trajectories.}
\label{tab:agent_repair_cells}
\setlength{\tabcolsep}{3.4pt}
\renewcommand{\arraystretch}{0.97}
\begin{tabular}{@{}lcccccccccccc@{}}
\toprule
\multicolumn{13}{@{}l}{\textit{Final accuracy} $\uparrow$} \\
\midrule
\multirow{2}{*}{Model} & \multicolumn{2}{c}{VQA-RAD} & \multicolumn{2}{c}{SLAKE} & \multicolumn{2}{c}{PathVQA} & \multicolumn{2}{c}{MedXpert} & \multicolumn{2}{c}{PMC-VQA} & \multicolumn{2}{c}{Average} \\
\cmidrule(lr){2-3} \cmidrule(lr){4-5} \cmidrule(lr){6-7} \cmidrule(lr){8-9} \cmidrule(lr){10-11} \cmidrule(lr){12-13}
 & Van. & MC & Van. & MC & Van. & MC & Van. & MC & Van. & MC & Van. & MC \\
\midrule
Qwen3-VL & 34.0 & 59.5 & 60.0 & 65.5 & 34.5 & 42.5 & 26.0 & 25.5 & 23.5 & 26.0 & \multirow{6}{*}{38.6} & \multirow{6}{*}{41.3} \\
Gemma-4 & 52.5 & 53.0 & 58.5 & 59.0 & 20.0 & 23.0 & 23.0 & 21.5 & 29.0 & 29.5 &  &  \\
Phi-4-MM & 33.5 & 46.0 & 48.0 & 49.0 & 35.0 & 34.5 & 19.0 & 23.0 & 17.0 & 18.0 &  &  \\
MedGemma & 56.0 & 55.5 & 60.5 & 61.5 & 35.0 & 36.0 & 24.5 & 25.0 & 22.5 & 22.0 &  &  \\
Lingshu & 56.0 & 57.0 & 70.0 & 70.0 & 47.0 & 55.5 & 18.0 & 19.0 & 26.5 & 31.0 &  &  \\
Huatuo & 65.5 & 65.5 & 60.5 & 63.5 & 47.5 & 48.5 & 20.5 & 19.5 & 34.0 & 33.0 &  &  \\
\midrule
\multicolumn{13}{@{}l}{\textit{CFA} $\downarrow$} \\
\midrule
\multirow{2}{*}{Model} & \multicolumn{2}{c}{VQA-RAD} & \multicolumn{2}{c}{SLAKE} & \multicolumn{2}{c}{PathVQA} & \multicolumn{2}{c}{MedXpert} & \multicolumn{2}{c}{PMC-VQA} & \multicolumn{2}{c}{Average} \\
\cmidrule(lr){2-3} \cmidrule(lr){4-5} \cmidrule(lr){6-7} \cmidrule(lr){8-9} \cmidrule(lr){10-11} \cmidrule(lr){12-13}
 & Van. & MC & Van. & MC & Van. & MC & Van. & MC & Van. & MC & Van. & MC \\
\midrule
Qwen3-VL & 95.1 & 13.4 & 95.7 & 24.6 & 95.7 & 21.6 & 86.6 & 1.3 & 99.3 & 24.5 & \multirow{6}{*}{87.7} & \multirow{6}{*}{42.7} \\
Gemma-4 & 100.0 & 43.0 & 97.6 & 42.2 & 97.5 & 21.7 & 87.3 & 63.9 & 99.3 & 46.4 &  &  \\
Phi-4-MM & 76.9 & 1.9 & 65.7 & 2.9 & 86.3 & 0.8 & 63.6 & 2.6 & 72.8 & 1.2 &  &  \\
MedGemma & 78.4 & 80.7 & 87.0 & 83.1 & 66.7 & 82.9 & 96.7 & 92.1 & 94.9 & 99.4 &  &  \\
Lingshu & 79.8 & 34.5 & 83.6 & 34.5 & 84.8 & 46.7 & 82.7 & 38.9 & 82.1 & 62.1 &  &  \\
Huatuo & 98.5 & 33.8 & 94.7 & 55.3 & 94.2 & 54.8 & 98.1 & 88.1 & 96.2 & 67.4 &  &  \\
\bottomrule
\end{tabular}
\end{table}

\newpage

\section{Prompts and Details}
\label{app:prompts_implementation}

\newtcolorbox{PromptCard}[1]{
  enhanced,
  colback=white,
  colframe=black!40,
  colbacktitle=black!7,
  coltitle=black,
  fonttitle=\bfseries\small,
  title=#1,
  boxrule=0.5pt,
  arc=1mm,
  left=7pt, right=7pt, top=5pt, bottom=5pt,
  before skip=7pt, after skip=7pt,
  breakable=false
}
\DefineVerbatimEnvironment{PromptText}{Verbatim}{
  fontsize=\footnotesize,
  breaklines=true,
  breakanywhere=false,
  breaksymbolleft={},
  breaksymbolright={}
}

\subsection{Generation Prompt and Verification Instructions}

\paragraph{Generation prompt.}
The common one-pass generator request uses the following user message.
\begin{PromptCard}{One-pass generator | user message}
\begin{PromptText}
You are a medical imaging expert. Look at the image carefully and answer the following question concisely.

Question: {question}

Answer:
\end{PromptText}
\end{PromptCard}

\begin{PromptCard}{Vanilla self-verification | user message}
\begin{PromptText}
You are a medical imaging verification expert. Determine whether the proposed answer to the medical imaging question is CORRECT or INCORRECT.

Question: {question}
Proposed Answer: {answer}

Respond in the following format:
Verdict: [CORRECT or INCORRECT]
Confidence: [0.0 to 1.0]
Explanation: [brief justification]
\end{PromptText}
\end{PromptCard}

\subsection{Actor-Verifier Loop Prompts}

\paragraph{The loop reuses the verifier verdict and explanation as feedback.}
Generation and feedback use the following templates.
\begin{PromptCard}{Agent-loop generator prompt}
\begin{PromptText}
You are a medical imaging expert. Look at the image carefully and answer the following question concisely.

Question: {question}

Answer:
\end{PromptText}
\end{PromptCard}

\begin{PromptCard}{Agent-loop verifier prompt}
\begin{PromptText}
You are a medical imaging verification expert. Determine whether the proposed answer to the medical imaging question is CORRECT or INCORRECT.

Question: {question}
Proposed Answer: {answer}

Respond in the following format:
Verdict: [CORRECT or INCORRECT]
Confidence: [0.0 to 1.0]
Explanation: [brief justification]
\end{PromptText}
\end{PromptCard}

\begin{PromptCard}{Agent-loop feedback regeneration prompt}
\begin{PromptText}
You are a medical imaging expert. You previously answered a medical imaging question, but a verifier assessed your answer as {VERDICT}.

Question: {question}

Your previous answer: {prev_answer}
Verifier's explanation: {explanation}

Please look at the image again carefully, reconsider, and provide a corrected answer.

Answer:
\end{PromptText}
\end{PromptCard}

When the verifier supplies no explanation, the feedback template omits the
explanation line; the verdict field is uppercased before insertion.

\clearpage
\subsection{Semantic-Judge Prompts}
\label{app:semantic_judge_prompts}

\paragraph{Primary semantic judge.}
The semantic judge is used only after inference. The user-message card below shows the answer-matching instructions in abbreviated form.
\begin{PromptCard}{Semantic answer judge | system message}
\begin{PromptText}
You are a strict medical VQA evaluator.
\end{PromptText}
\end{PromptCard}

\begin{PromptCard}{Semantic answer judge | user message}
\begin{PromptText}
You are evaluating medical visual question answering outputs.
Given the question, the reference answer, and the model answer, decide whether the model answer should be counted as correct.

Rules:
- Return CORRECT only if the model answer semantically matches the reference answer.
- Accept clear synonyms or equivalent medical phrasing.
- Do not give partial credit.
- Return only CORRECT or INCORRECT.

Question: {question}
Reference Answer: {reference_answer}
Model Answer: {model_answer}
\end{PromptText}
\end{PromptCard}



\subsection{Limitations and Broader Impact}

\paragraph{Limitations.}
\ours{} incurs additional inference cost because it separates contrast construction, visual measurement, and evidence-to-verdict binding into multiple model calls. We study this decomposition to isolate the reliability benefit of evidence-bound verification rather than to optimize latency or serving cost. Practical deployment may therefore require more efficient implementations. Our experiments also focus on benchmark questions whose answers can be assessed from the available image. The approach does not address cases whose correctness depends primarily on external clinical history, laboratory measurements, or other information absent from the image. Finally, we use a validated semantic-judging procedure with human audit to reduce this source of evaluation error, although residual ambiguity in benchmark answers cannot be eliminated entirely.

\paragraph{Broader impact.}
This work is intended to support model risk assessment and the design of more reliable verification mechanisms. Our experiments evaluate controlled medical VQA rather than end-to-end clinical decision-making. The results should not be
interpreted as validating autonomous clinical use. Deployment in clinical settings would require task-specific external validation, calibrated uncertainty, and appropriate human oversight.